\documentclass[conference]{IEEEtran}
\IEEEoverridecommandlockouts
\usepackage[hidelinks]{hyperref}
\usepackage{cite}
\usepackage{amsmath,amssymb,amsfonts}
\usepackage{algorithmic}
\usepackage{graphicx}
\usepackage{textcomp}
\usepackage{xcolor}
\usepackage[linesnumbered,ruled,vlined]{algorithm2e}
\usepackage{balance}
\usepackage{multirow}
\usepackage{enumitem}
\usepackage{soul}
\usepackage{tikz}
\usepackage{diagbox}
\usepackage{calc}
\usepackage{makecell}
\usepackage{amsmath}
\usepackage{booktabs}
\usepackage{subfigure}
\usepackage{tikz}
\usepackage{framed}
\usepackage{lipsum}
\usepackage{color}
\definecolor{shadecolor}{rgb}{0.92,0.92,0.92}

\newcommand*\wcircled[1]{\tikz[baseline=(char.base)]{\node[shape=circle,draw,inner sep=0.75pt] (char) {\textcolor{black}{#1}};}}
\newcommand*\scircled[1]{\tikz[baseline=(char.base)]{\node[shape=circle,fill,inner sep=0.75pt] (char) {\textcolor{white}{#1}};}}

\newcommand{\add}[1]{\textcolor{black}{{#1}}\xspace}

\newcommand{\annotate}[1]{\textcolor{gray}{{#1}}\xspace}

\newcommand{\trivia}{\texttt{TRIVIA\_QA}\xspace}

\def\BibTeX{{\rm B\kern-.05em{\sc i\kern-.025em b}\kern-.08em
    T\kern-.1667em\lower.7ex\hbox{E}\kern-.125emX}}

\newcommand{\our}{\texttt{SAGE}\xspace}
\newcommand{\oursys}{\texttt{SAGE}\xspace}
\begin{document}

\title{\oursys: A Framework of Precise Retrieval for RAG}

\author{
\IEEEauthorblockN{Jintao Zhang}
\IEEEauthorblockA{\textit{Department of Computer Science}\\
\textit{Tsinghua University}\\
zhang-jt24@mails.tsinghua.edu.cn}
\and
\IEEEauthorblockN{Guoliang Li\textsuperscript{*}}
\IEEEauthorblockA{\textit{Department of Computer Science}\\
\textit{Tsinghua University}\\
liguoliang@tsinghua.edu.cn}
\and
\IEEEauthorblockN{Jinyang Su}
\IEEEauthorblockA{\textit{Department of Computer Science}\\
\textit{Tsinghua University}\\
sujinyanslip@gmail.com}

\thanks{Guoliang Li is the corresponding author. This paper was supported by National Key R\&D Program of China (2023YFB4503600), NSF of China (61925205, 62232009, 62102215), Huawei, TAL education, and Beijing National Research Center for Information Science and Technology (BNRist).}
}

\maketitle

\begin{abstract}

Retrieval-augmented generation (RAG) has demonstrated significant proficiency in conducting question-answering (QA) tasks within a specified corpus. Nonetheless, numerous failure instances of RAG in QA still exist. These failures are not solely attributable to the limitations of Large Language Models (LLMs); instead, they predominantly arise from the retrieval of inaccurate information for LLMs due to two limitations: (1) Current RAG methods segment the corpus without considering semantics, making it difficult to find relevant context due to impaired correlation between questions and the segments. (2) There’s a trade-off between missing essential context with fewer context retrieved and getting irrelevant context with more context retrieved. It is hard to make an ideal balance.

In this paper, we introduce a RAG framework, named \oursys, designed to overcome these limitations. First, to address the issue of segmentation without considering semantics, we propose to train a semantic segmentation model. This model is trained to segment the corpus into semantically complete chunks. Second, to ensure that only the most relevant chunks are retrieved while the irrelevant ones are ignored, we design a chunk selection algorithm to dynamically select chunks based on the decreasing speed of the relevance score of chunks, leading to a more relevant selection.
Third, to further ensure the precision of the retrieved chunks, we propose letting LLMs assess whether retrieved chunks are excessive or lacking and then adjust the amount of context accordingly.
Experimental results show that \oursys outperforms baselines by 61.25\% in the quality of QA on average. Moreover, by avoiding retrieving noisy context, \oursys lowers the cost of the tokens consumed in LLM inference and achieves a 49.41\% enhancement in cost efficiency on average. Additionally, our work offers valuable insights for boosting RAG, contributing to the development of more effective RAG systems.

\end{abstract}

\section{Introduction}  \label{sec:intro}
\setcounter{figure}{0}
\setcounter{table}{0}

Retrieval-augmented generation (RAG) is a technique that enhances a generation model's ability to answer questions for a given corpus by retrieving information related to the question. With the rise and advancement of large language models, RAG has demonstrated remarkable proficiency in QA tasks across both commercial applications and open-source communities~\cite{edge2024local,llamaindex,sarthi2024raptor}. 

\noindent\textbf{Limitations.} Typically, a RAG system operates in three distinct phases. \underline{First}, the given \textbf{corpus} will be segmented into many \textbf{chunks}. \underline{Second}, in response to a specific question, a retriever identifies and selects the top $K$ most related chunks to use as \textbf{context}. \underline{Third}, the question, alongside the context, will be inputted into a LLM to generate an answer. 
Therefore, the effectiveness of a RAG system heavily relies on three pivotal components: an effective and efficient method for segmenting the corpus into chunks in the first stage, an accurate mechanism for retrieving the most relevant chunks in the second stage, and, finally, a LLM proficient in understanding and processing natural language for question answering in the third stage.
Apart from the limitations of LLMs, current RAG systems have the following critical limitations. 
% \lgl{first talk about the workflow of RAG and then give the limitations of each step.}

\begin{figure}[!t]
  \centering
  \vspace{.1em}
  \includegraphics[width=0.49\textwidth]{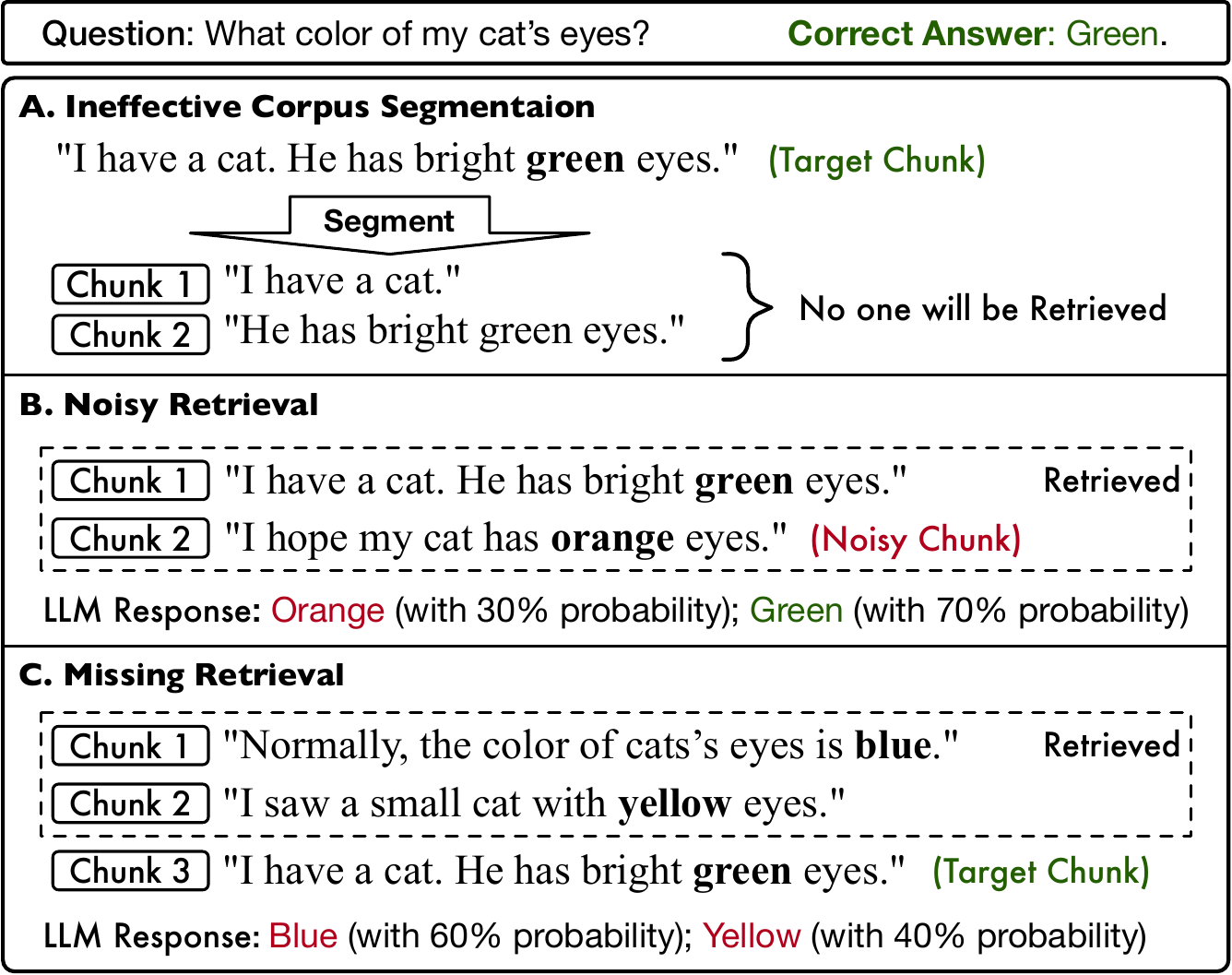}
  \vspace{-1.2em}
  \caption{Three motivational examples illustrating the current limitations of precise retrieval for RAG.}
  \vspace{-1em}
  \label{fig:intro_limitation}
\end{figure}

\noindent\textbf{(L1)} \textbf{Ineffective Corpus Segmentation:} Often, RAG systems segment the corpus into fixed-length chunks~\cite{teja2023evaluating, sarthi2024raptor} without effectively considering semantic coherence~\cite{gao2023retrieval}. Consequently, retrieved chunks convey incomplete meanings, leading to incorrect answers.
For example, Figure~\ref{fig:intro_limitation} (A) illustrates a scenario where semantic-based segmentation is not applied. The \texttt{Target Chunk} (the segment crucial for deriving the correct answer) is segmented into two parts. Such segmentation can make these segments semantically unrelated to the question \textit{“What is the color of my cat's eyes?"} when assessed independently. Consequently, the probability of retrieving both necessary segments diminishes, leaving the LLM unable to provide the correct answer without the full \texttt{Target Context}.

\noindent \textbf{Challenge of addressing (L1):} Actually, with the advanced natural language understanding capabilities of LLMs, such as \texttt{GPT-4}~\cite{achiam2023gpt}, there is a good approach to do segmentation. We can input the entire corpus along with a segmentation command (\texttt{"Please segment the corpus into chunks semantic completely."}) into a LLM, and get the chunks. However, such a method is impractical due to its high costs and prolonged processing time. For instance, segmenting a corpus of $1e6$ tokens with \texttt{GPT-4} could spend more than $90$ dollars and about $8$ hours to complete. Such requirements are often unrealistic for most applications. We need to design a much quicker and more cost-effective solution.

\noindent\textbf{(L2)} \textbf{Noisy and Missing Retrieval:} Current RAG systems retrieve the top $K$ chunks deemed most relevant to the given question. However, this approach often leads to two significant issues: 
\underline{(1)} \textit{Noisy Retrieval}: This refers to instances where irrelevant information, called \texttt{Noisy Chunks}, are retrieved alongside relevant ones, misleading the LLM to produce incorrect responses~\cite{cuconasu2024power}. Such chunks are unhelpful for answering questions. The problem arises because systems aim to avoid overlooking potentially useful information by retrieving a fixed, and often excessive, number ($K$) of chunks as context for the LLM. For example, as demonstrated in Figure~\ref{fig:intro_limitation} (B), retrieving two chunks instead of one might result in a 30\% chance that the LLM will incorrectly answer \texttt{Orange}. Here, the first chunk is the \texttt{Target Chunk}, and the second one is a \texttt{Nosiy Chunk}. In this case, Setting $K$ to 1 could allow the LLM to give the correct answer, but the $K$ is set to $2$ because it is hard to ensure that the \texttt{Target Chunk} will always be ranked first by the retriever. Such a strategy will easily retrieve misleading information that prevents the LLM from getting the correct answer. 
\underline{(2)} \textit{Missing Retrieval}: This issue occurs when the \texttt{Target Chunk} is not among the retrieved chunks, thereby losing crucial context. For example, Figure~\ref{fig:intro_limitation} (C) shows a scenario where the retriever ranks the \texttt{Target Chunk} third while $K$ is set to 2. Then the \texttt{Target Chunk} is missing in the context, eliminating any chance of a correct response. This issue is because ensuring the \texttt{Target Chunk} ranks within the top $K$ chunks is difficult for retrievers~\cite{cuconasu2024power}.

\noindent \textbf{Challenge of addressing (L2):} A seemingly straightforward method is selecting the optimal fixed value for $K$. However, the trade-off between \textit{Noisy Retrieval} and \textit{Missing Retrieval} always exists for a fixed $K$. Specifically, setting a larger $K$ can increase the likelihood of incorporating \texttt{Noisy Chunks}, leading to more errors from \textit{Noisy Retrieval}. Conversely, setting a small $K$ risks losing \texttt{Target Chunk}, resulting in inaccuracies due to \textit{Missing Retrieval}. It is necessary to devise a method to determine the most appropriate value for $K$ dynamically, balancing the need to minimize both types of retrieval errors.

\noindent\textbf{Our approach.} To overcome these limitations, we develop a novel RAG framework, named \oursys,  which incorporates \textit{\textbf{s}emantic segment\textbf{a}tion}, \textit{\textbf{g}radient-based chunk selection}, and \textit{self-f\textbf{e}edback of LLMs} to facilitate precise retrieval for RAG. \oursys is designed to tackle specific limitations as follows: To overcome \textbf{(L1)}, we propose to train a lightweight model to rapidly and accurately segment the corpus into semantically coherent chunks, ensuring that the retrieved information is semantically complete and relevant. Moreover, because our segmentation method divides the corpus into the smallest segments with complete semantics, it can minimize the number of context tokens required, thereby lowering the inference cost for the LLM in a RAG system.
To overcome \textbf{(L2)}, we propose to select the most relevant chunks dynamically. Instead of retrieving a fixed number of the top $K$ chunks, we employ a sophisticated model to score each chunk, arranging them in descending order of relevance. We then select the most relevant chunks up to the point where a significant drop in relevance scores occurs. This method prioritizes highly relevant chunks, preventing \texttt{Noisy Chunks} from being fed into the LLM.
Additionally, to further ensure that the final context contains \texttt{Target Chunk} while excluding \texttt{Noisy Chunks}, we integrate a self-feedback mechanism. Specifically, this technique leverages a LLM to assess if the retrieved chunks are excessive or insufficient for accurate QA. Based on this assessment, the amount of chunks to be retrieved is adjusted automatically.

By overcoming these limitations, \oursys enhances RAG systems' capability to retrieve precise context, thereby facilitating the generation of accurate answers. Moreover, through the elimination of semantically incomplete and noisy chunks, \oursys lowers the cost of the tokens consumed during LLM inference.

\begin{figure*}[!t]
\centering
\includegraphics[width=0.775\textwidth]{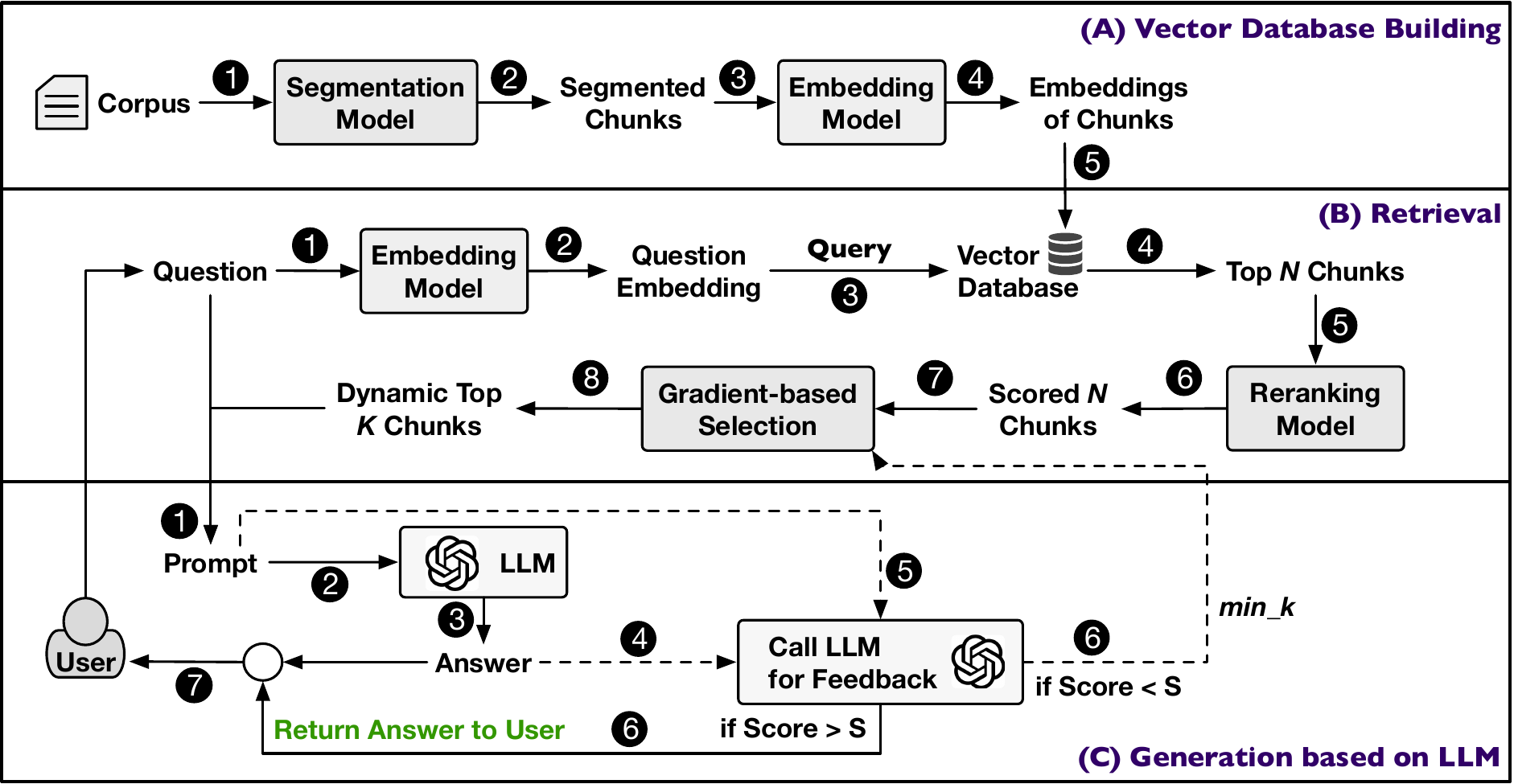}
\vspace{-1em}
\caption{Workflow of \oursys, where the $\dashrightarrow$ inidcates the pipelines of self-feedback.}
\label{fig:overview}
\end{figure*}

\noindent\textbf{Contributions.}  
Our key contributions are summarized below.

\noindent\textbf{(C1)} We propose a semantic segmentation method that segments a corpus into short, semantically coherent chunks quickly, improving the QA capabilities of RAG.

\noindent\textbf{(C2)} We develop a gradient-based chunk selection method that dynamically selects the most relevant chunks while eliminating irrelevant chunks for RAG.

\noindent\textbf{(C3)} We implement a self-feedback mechanism to adjust the number of retrieved chunks, further ensuring the precision of retrieval. 

\noindent\textbf{(C4)} Through detailed experimentation, we demonstrate that our RAG framework outperforms existing baselines in both QA ability and cost-efficiency.

\noindent\textbf{(C5)} We offer valuable insights into RAG tasks, providing researchers in the field with guidance for developing more effective RAG systems.
\begin{table}[!h]
    \centering
    \caption{Notations.}
    \vspace{-1.2em}
    \label{tab:notations}
    \setlength\tabcolsep{9pt}
        \begin{tabular}{|c|c|} \hline
        \textbf{Notation}  &  \textbf{Description} \\ \hline
        % $O$  &   Corpus   \\ \hline
        $\mathbb{T}$  &  Segmented Chunks  \\ \hline
        
        $f_e(\cdot)$   &   Embedding model    \\ \hline
        $\mathcal{M}$    &   MLP model used in segmentation model    \\ \hline
        % $f_r(\cdot)$   &   Reranking model    \\ \hline
        $\mathbb{C}$     &   Chunks queried from vector database    \\ \hline
        $\mathbb{C}_s$   &   Chunks after gradient based selection    \\ \hline
        $N$     &   The number of chunks queried from a vector database  \\ \hline
        $K$     &   The number of retrieved chunks  \\ \hline
        $c_i, c_o$     &   Price per input/output token of a LLM    \\ \hline
        % $t$     &   inference latency per token of a LLM    \\ \hline
        \end{tabular}
        \vspace{-1em}
\end{table}

\section{Preliminaries}

\subsection{Retrieval-augmented generation (RAG)}

A RAG system operates through three principal phases:

\textbf{(1)} Vector Database Construction: Initially, the selected corpus is divided into segments, or "\textbf{chunks}," which are then converted into vector representations using an embedding model. These vector representations are subsequently stored in a vector database for later retrieval.

\textbf{(2)} Retrieval: Upon receiving a question, the same embedding model used in the previous phase converts this question into a vector. This question vector then serves to perform a query process within the vector database, identifying the top $N$ chunks' vectors most similar to it, typically determined through the shortest cosine distance. These $N$ chunks are then extracted as the context for a LLM.

\textbf{(3)} Answer Generation: The given question alongside the retrieved chunks is arranged into an appropriate prompt. This prompt is then fed into an LLM, which generates a response as the final answer to the question.

Through this structured approach, the RAG system could retrieve relevant context from
the corpus to enhance the accuracy of answering the question.

\subsection{Cost of LLM inference}

The cost of LLM inference could be quantified by examining how much money is required to obtain answers from a LLM. Typically, many RAG systems utilize services from LLM providers, which incur charges based on the volume of input and output tokens processed by a designated LLM. For instance, OpenAI's GPT-4~\cite{achiam2023gpt} might charge 10 dollars for every one million (1e6) input tokens and 30 dollars for the same quantity of output tokens, respectively. 
Therefore, the inference cost of a LLM is determined by calculating the expenses incurred for both input and output tokens processed by the LLM as follows.

\begin{equation}
\label{equ:LLM cost} 
Cost = I_t * c_i + O_t * c_o
\end{equation}

Where $I_t$ and $O_t$ indicate the number of input tokens and output tokens of a LLM, respectively. $c_i$ and $c_o$ mean the cost per input token and per output token of the LLM, respectively.

\subsection{Cost efficiency Metric in RAG}

We introduce a cost efficiency metric that considers both the quality of QA and the cost associated with LLM inference. The equation is given as follows:

\begin{equation}
\label{equ:cost efficiency} 
Cost-efficiency = \frac{Acc}{Cost}
\end{equation}

In the above equation, $Acc$ means accuracy or any other metrics used to evaluate the quality of an answer for a given question, such as the F1-Score~\cite{rajpurkar2016squad} or BELU-1~\cite{papineni2002bleu}. Meanwhile, $Cost$ is derived from the cost Equation~\ref{equ:LLM cost}. Higher cost efficiency implies a better performance-to-cost ratio.

\section{Overview}  \label{sec:overview}
We will introduce the workflow of \oursys as shown in Figure~\ref{fig:overview}.

\subsection{Vector Database Creation}

As illustrated in Figure~\ref{fig:overview} (A), \scircled{1}-\scircled{2}, we first employ a trained segmentation model to segment each paragraph split by \verb|'\n'| in a corpus into short but semantically complete chunks, arranging them is a set denoted as $\mathbb{T}$.
Following segmentation, \scircled{3}-\scircled{4} we apply an embedding model, represented as $f_e(\cdot)$, to convert the chunks $\mathbb{T}$ into a collection of vector embeddings $f_e(\mathbb{T})$. \scircled{5} These embeddings are then stored in a vector database. Importantly, we maintain a record of the mapping between the index of each chunk in $\mathbb{T}$ and its corresponding vector in $f_e(\mathbb{T})$. This allows us to retrieve a particular chunk based on its vector embedding easily.

\subsection{Retrieval}

As illustrated in Figure~\ref{fig:overview} (B), our retrieval process is designed to identify chunks that assist in answering a specific question. Initially, \scircled{1}-\scircled{2} the question received from a user is transformed into a vector through the embedding model $f_e(\cdot)$. Following this, \scircled{3}-\scircled{4} we proceed to query the vector database to extract the $N$ vectors that most closely align with the question's embedding, thereby retrieving the corresponding $N$ chunks. Subsequently, \scircled{5}-\scircled{6} these $N$ chunks undergo evaluation by a reranking model, which assigns scores based on their relevance to the question. Finally, \scircled{7}-\scircled{8}, we select the top-ranking $K$ chunks that come before the point where a significant dip in scores is observed, ensuring that only the chunks most related to the question are chosen for the next phase.
\add{We can regard the retrieval process as two stages. The first stage involves a simple process of querying a vector database, while the second stage scores the retrieved chunks using a more complex model. Relying solely on the retriever may not efficiently rank results, while using only a reranker can result in high latency. 
This two-stage recall approach is a common technique in real-world information retrieval systems.}

\subsection{Generation}

As demonstrated in Figure~\ref{fig:overview} (C), we integrate the retrieved chunks and the posed question into a LLM to generate an answer for the user. Specifically, \scircled{1} we craft a prompt incorporating both the question and the retrieved chunks, tailored to the question's type—be it multiple-choice or open-ended. \scircled{2}-\scircled{3} This prepared prompt is then inputted into an LLM to procure the response.
Crucially, \scircled{4}-\scircled{5} we further organize the generated answer alongside the initial question and the retrieved chunks into a feedback prompt. The purpose of this feedback prompt is twofold: 1) to evaluate the quality of the answer and 2) to assess whether the selected chunks are whether excessive
or insufficient for accurate QA. By submitting the feedback prompt to an LLM, we acquire both the score of the answer and an assessment of the chunks.
If the answer's score surpasses a predetermined threshold, \scircled{6}($\dashrightarrow$)-\scircled{7} it is subsequently returned to the user. Conversely, \scircled{6}($\rightarrow$) adjustments are made to the value of $K$ based on the chunks' assessment. Following this, \scircled{6}($\dashrightarrow$) we go through the gradient-based chunk selection and generation processes again to improve the answer until the score of the answer surpasses the threshold or the feedback loop has been executed three times.

\section{Semantic Segmentation} \label{sec:segmentation}

\begin{figure}[!h]
\centering
\includegraphics[width=0.49\textwidth]{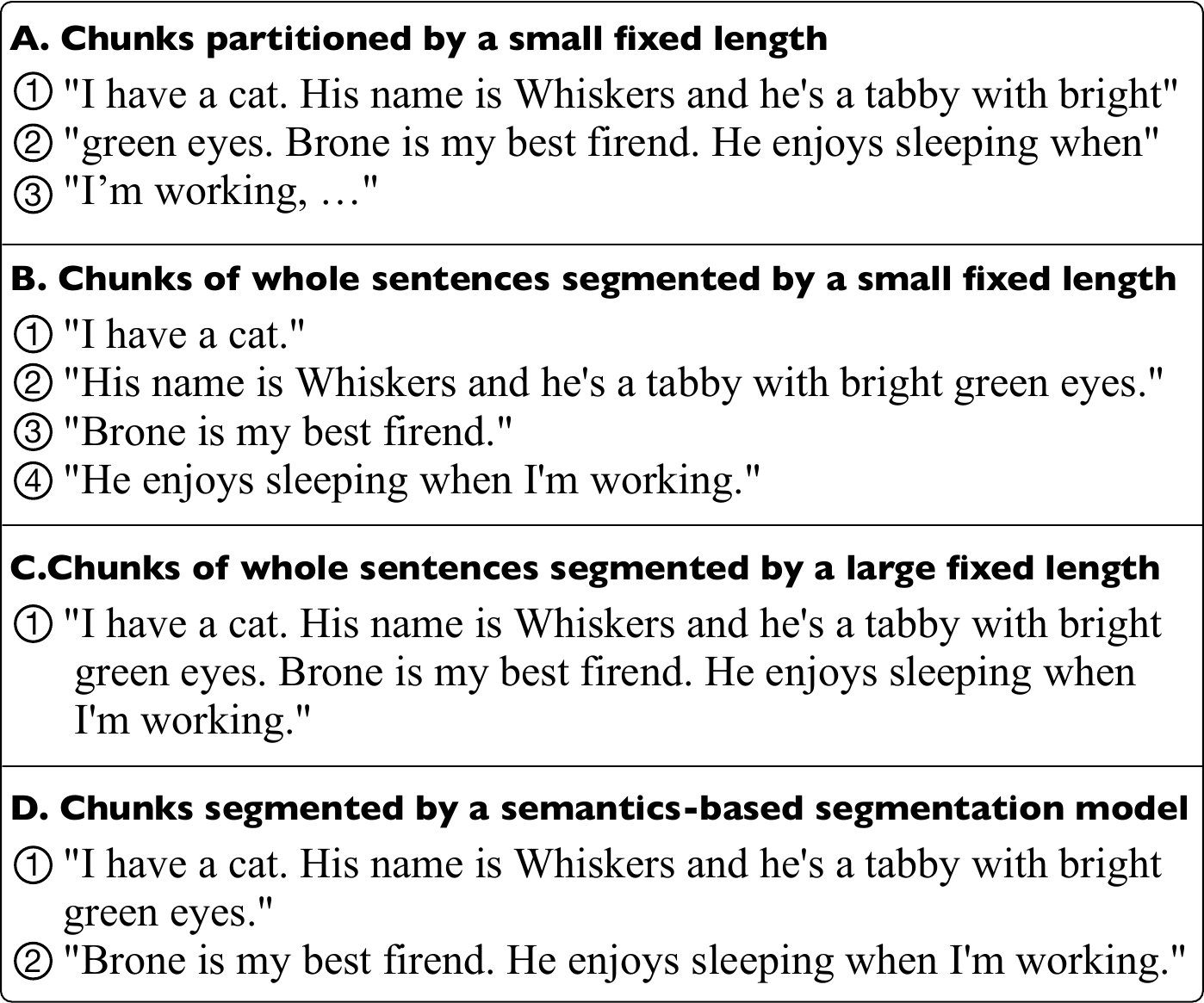}
\vspace{-1.5em}
\caption{Motivation of corpus segmentation. The number in \protect\wcircled{\textcolor{white}{1}} means the chunk ID.}
\vspace{-1em}
\label{fig:partition_motivation}
\end{figure}

\begin{figure}[!h]
    \centering
    \includegraphics[width=0.495\textwidth]{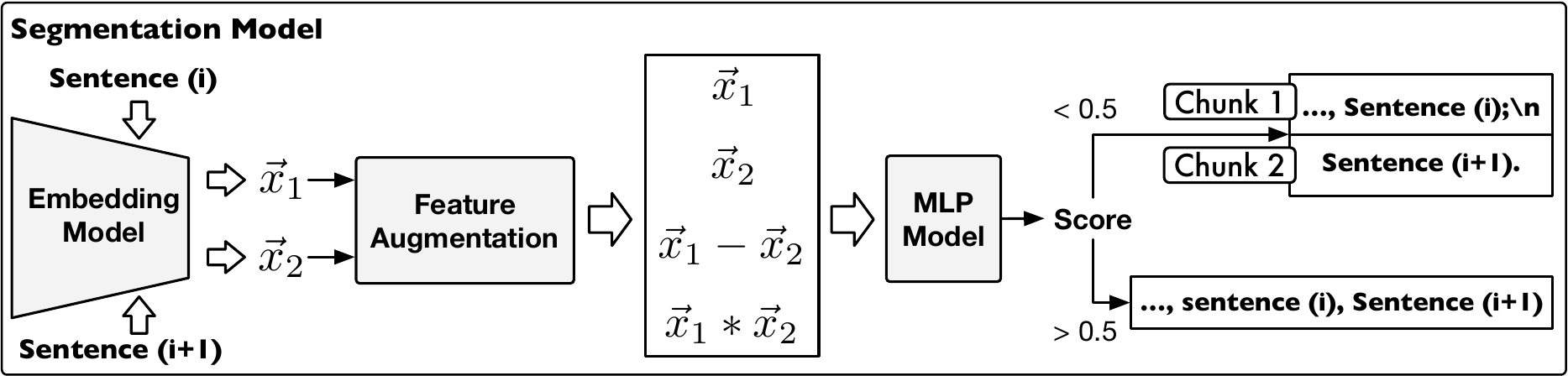}
    \vspace{-1.5em}
    \caption{Corpus segmentation model.}
    \vspace{-1em}
    % \vspace{-.15em}
    \label{fig:partition_model}
\end{figure}

\subsection{High-level Idea}
Corpus segmentation plays a crucial role in a RAG system. Ineffective segmentation can often result in semantically incomplete chunks, leading to the retrieval of irrelevant and incomplete information~\cite{gao2023retrieval}, resulting in incorrect answers. We also show this observation with an experimental case presented in Section~\ref{sec:insight}.

At present, two segmentation methods are predominantly employed in RAG systems. The first method divides the corpus into segments based on a predetermined number of tokens. The second method builds upon the first method by ensuring that each chunk contains complete sentences. Both methods frequently fail to produce semantically complete chunks.
Specifically, the first, more straightforward strategy involves segmenting the corpus based on a predetermined number of tokens. This method often leads to chunks containing incomplete sentences, thereby undermining the overall coherence and meaning of each chunk. Figure~\ref{fig:partition_motivation}-A illustrates this issue, displaying chunks that are filled with incomplete sentences. Consequently, calculating the similarity between a user's question and these chunks becomes ineffective. 
Another widely used strategy involves segmenting contiguous sentences less than a fixed length into a chunk. According to this approach, if a chunk exceeds the predetermined length limit, the last sentence is transferred intact to the following chunk instead of being truncated mid-sentence. This ensures that each chunk comprises complete sentences. However, this method can also distort the meaning of a chunk. As depicted in Figure~\ref{fig:partition_motivation}-B, the first and second chunks are concatenated in the original corpus. Once separated from the first chunk, the pronouns \verb|'His'| and  \verb|'He'| in the second chunk become unclear references. Consequently, computing the similarity between a user's question and such chunks is also problematic because these chunks lack coherence, impairing the representativeness of their embeddings as well.
Increasing the fixed length, as demonstrated in Figure~\ref{fig:partition_motivation}-C, can mitigate the issue of chunks having incomplete meanings. However, this adjustment may cause each chunk to contain an excessive amount of information. If a user's question targets a specific segment of the corpus, the similarity calculation between the question and these overloaded chunks may be unsuccessful. Furthermore, even retrieving these overloaded chunks, this approach can result in feeding an excessive number of irrelevant tokens into the LLM, interfering with the quality of QA and significantly increasing costs.
Note that while employing a LLM for segmentation is a conceivable approach, it proves to be costly (See Section~\ref{sec:exp_cost}). This is because the LLM requires the inputting and outputting of all tokens of the given corpus. Additionally, this method is markedly slow, as the LLM's capacity for parallel processing is limited by its substantial GPU memory and computation requirements.

To address these issues, we propose to develop a lightweight and effective segmentation model. As illustrated in Figure~\ref{fig:partition_motivation}-D, chunks segmented by our model ensure that each chunk conveys a focused and complete meaning without containing an excessive number of tokens.

\subsection{Model Construction}

As demonstrated in Figure~\ref{fig:partition_model}, our segmentation model employs a structure that integrates an embedding model with a Multi-layer Perceptron (MLP) model.

The model is structured into three main components. Initially, the embedding model, which utilizes a state-of-the-art and lightweight design~\cite{li2023angle}, embeds two sentences to generate two vectors, denoted as $x_1$ and $x_2$. Subsequently, a feature augmentation module receives $x_1$ and $x_2$, performing operations to subtract and multiply these embeddings, yielding the results of their difference $x_1 - x_2$ and product $x_1 * x_2$. The final component, an MLP model, takes both the $x_1$ and $x_2$, alongside ($x_1 - x_2$) and ($x_1 * x_2$), to produce a score. This score determines whether the two input sentences should be segmented or kept as a contiguous part.
\add{The decision to include both ($x_1 - x_2$) and ($x_1 * x_2$) in our model was based on the observation that even with context-sensitive embeddings like BERT, embeddings can still meaningfully reflect semantic differences or similarities between sentences~\cite{reimers2019sentence}.}

\setlength{\textfloatsep}{5pt}
\begin{algorithm}[!t]  
    \small
    \caption{Training of the segmentation model}
    \label{alg:segmentation training} 
    \KwIn{Some passages of WikiPedia $\mathbb{D}$, a embedding model $f_e(\cdot)$ and a MLP model $\mathcal{M}$.}
    % \KwOut{Trained MLP model $\mathcal{M}$.}
    Collect $N$ sentence pairs $\mathbb{P} = \{<s_1, s_2, label>_i\}$ from $\mathbb{D}$

    \For {each epoch in the training process} { 
        \For {$S$ \textbf{in} $\mathbb{S}$} { 
            $s_1, s_2, label = S$;

            $\vec x_1, \vec x_2 = f_e(s_1, s_2)$;

            $Score = \mathcal{M}(\vec x_1, \vec x_2, (\vec x_1 - \vec x_2), (\vec x_1 * \vec x_2))$;

            $Loss = MSE(Score, label)$; \tcp{\annotate{Mean squared error loss.}}

            Update $f_e(\cdot), \mathcal{M}$ according to $Loss$.;
        }
    }
    % \textbf{return} $\mathcal{M}$;
\end{algorithm}

\subsection{Model Training}

To develop a segmentation model capable of judging whether two sentences are closely related, it is essential to first gather a substantial amount of semantically segmented training data.
A good source for this is the Wikipedia dataset ~\cite{wikidump}, where almost all passages have been segmented semantically into paragraphs. Typically, sentences that are closely related appear within the same paragraph consecutively, whereas unrelated sentences are found in separate paragraphs. This structure allows for the collection of numerous sentence pairs, each pair comprising two sentences. These pairs are accompanied by a $label$ that indicates whether the sentences should be grouped into a single chunk.
Here, if two sentences are consecutive and within the same paragraph, then $lable = 1$, representing they should be grouped into the same chunk. Otherwise, $lable = 0$, suggesting they should be segmented into different chunks. 

The training procedures are detailed in Algorithm~\ref{alg:segmentation training}. We feed pairs of sentences from the collected dataset into the model sequentially, and learn the parameters of the embedding model and MLP model by adjusting them to fit the output score to the $label$ of each sentence pair, utilizing the gradient descent optimization method.

\subsection{Model Inference} \label{sec:segmentation inference}

The inference process of the segmentation model is straightforward. Each two adjacent sentences in the corpus is input into the segmentation model to obtain a score. If this score falls below a predetermined segmentation score threshold, $ss$, which ranges between 0 and 1, e.g., 0.5, the sentences are segmented into separate chunks. Conversely, if the score is above or equal to $ss$, the sentences are retained within the same chunk.

Note that the segmentation process for any given corpus is executed swiftly (See Section~\ref{sec:exp_cost}), because our lightweight segmentation model can be run in parallel in a GPU. For instance, we can gather all pairs of sentences within a corpus and organize them into multiple batches, each with a size of 512. Subsequently, the segmentation model is called to perform inference in parallel.

\subsection{Corpus Segmentation}  \label{subsec:corpus segment}
Given a corpus, we initially segment it into coarse-grained chunks, each comprising complete sentences, as depicted in Figure~\ref{fig:partition_motivation} (C). Subsequently, we employ our trained segmentation model further to segment these chunks into semantically coherent, fine-grained segments. Specifically, we begin by partitioning the corpus into chunks of approximately $l$ tokens in length. Then, our segmentation model evaluates each pair of adjacent sentences in these coarse-grained chunks, as described in Section~\ref{sec:segmentation inference}, getting the final fine-grained segments.

\begin{figure}[h]
    \centering
    \includegraphics[width=0.425\textwidth]{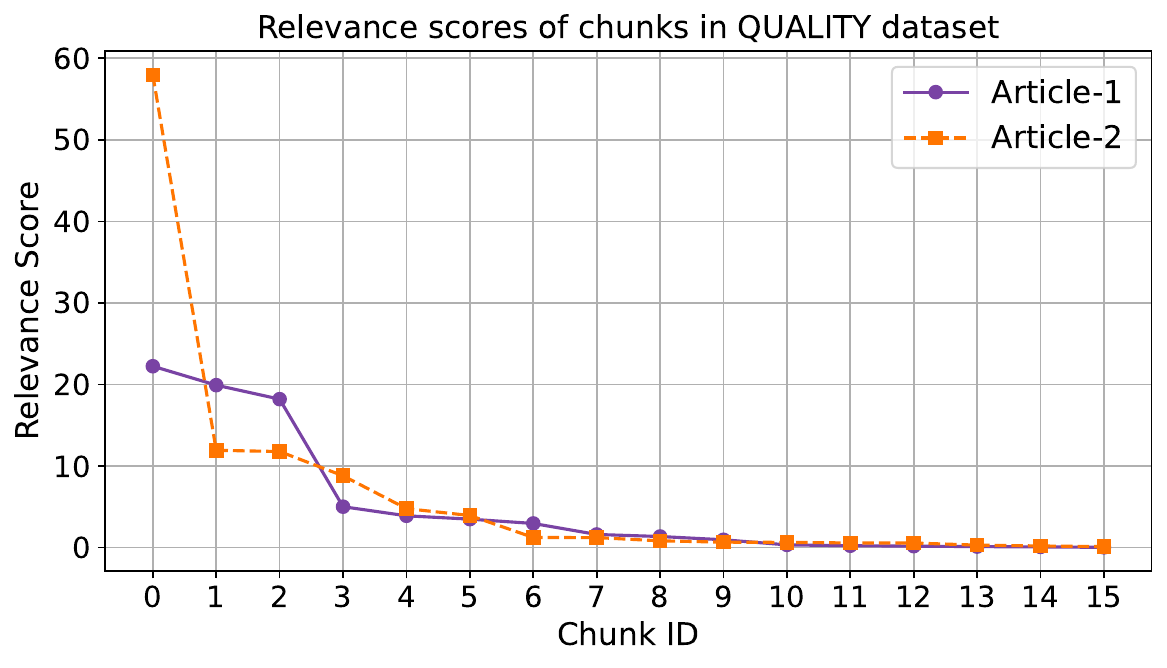}
    \vspace{-1em}
    \caption{Two general cases in relevance scores of retrieved segmentations.}
    \label{fig:reranking_motivation}
\end{figure}

\section{Gradient-based Chunk Selection} \label{sec:gradient selection}

\subsection{High-level Idea}

Advanced RAG systems typically leverage a reranking model to assess the relevance of chunks to a given question. The process involves using the reranking model to assign relevance scores to $N$ chunks queried from the vector database, which possesses the shortest embedding distance to the question. Subsequently, the $K$ highest-scoring chunks are selected as context for QA. To minimize the risk of overlooking useful chunks, $K$ is often set to a large number. However, such an approach can unintentionally include irrelevant chunks. Unnecessary information in the collected chunks can confuse LLMs, making it harder for them to give correct answers. This issue will be further illustrated through some experimental cases in Section~\ref{sec:insight}.
Fortunately, we observe that current state-of-the-art reranking models have the capability to precisely score chunks related to a given question, effectively assigning lower scores to the irrelevant ones. Furthermore, Figure~\ref{fig:reranking_motivation} shows two general cases of the relevance score of chunks for two articles in a dataset. The scoring pattern across chunks often reveals a sharp decline before a gradual slope. If we only select the top three chunks for Article-1 and one chunk for Article-2, the correct answer is easy to get. We will demonstrate specific cases in Section~\ref{sec:insight}. This indicates that chunks preceding the sharp drop are more significantly related to the question than those following it. Building on this observation, we propose a dynamic selection of chunks based on the gradient of the scores of sorted chunks, rather than sticking to a fixed number. We aim to identify the most relevant chunks to a given question more accurately.

\setlength{\textfloatsep}{3pt}
\begin{algorithm}[h]  
    \small
    \caption{Gradient-based Chunk Selection}
    \label{alg:Reranking} 
    \KwIn{$K$ chunks $\mathbb{C}$, Reranking Model $\mathcal{R}$, Retrieval minimum number $min\_k$, threshold of gradient $g$.}
    \KwOut{Retrieved chunks.}

    $\mathbb{S} = \mathcal{R(\mathbb{C})}$;

    Sort($\mathbb{C}, \mathbb{S}$);  \tcp{\annotate{Sort chunks based on scores}}

    $\mathbb{C}_s = \mathbb{C}[:, k]$;

    $score = \mathbb{C}[k-1]$;

    \For {each i in [$min\_k$, $N$)} { 
        \If{$\mathbb{S}[i] > score / g$} { 
            $\mathbb{C}_s.append(\mathbb{C}[i])$;
        }
        \Else{Break;}
    }
    \textbf{return} $\mathbb{C}_s$;
\end{algorithm}

\begin{figure*}[!t]
    \centering
    \includegraphics[width=0.93\textwidth]{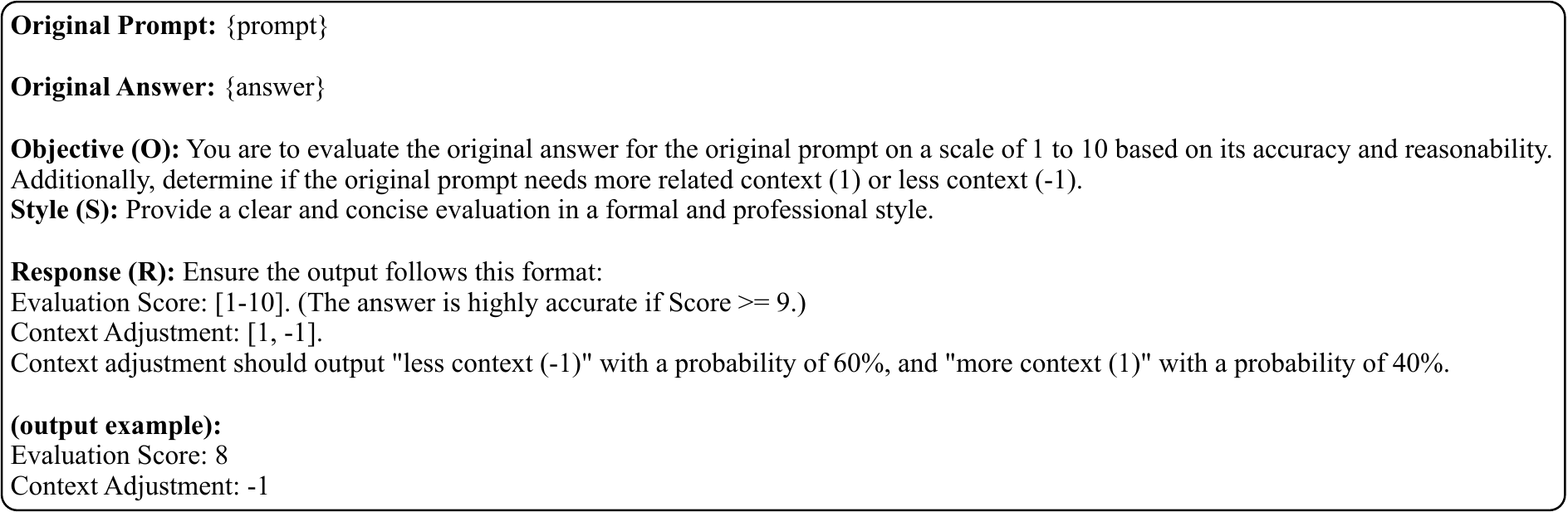}
    \vspace{-1em}
    \caption{Prompt of Self-Feedback.}
    \vspace{-.5em}
    \label{fig:feedback}
\end{figure*}

\subsection{Gradient based Selection} \label{subsec:gradient select}

Our chunk selection algorithm is outlined in Algorithm~\ref{alg:Reranking} and corresponds to steps \scircled{5}-\scircled{8} in Figure~\ref{fig:overview} (B). This algorithm requires three inputs. The first is $\mathbb{C}$, a collection of $N$ chunks that are closest in embedding distance to the question. The second input is $min\_k$, which specifies the minimum number of chunks the algorithm should return. The third input is a gradient threshold $g$ utilized to identify the significant drop in scores. We aim to select top chunks before a decrease rate of $g$ among the scored chunks in descending order. The output is $\mathbb{C}_s$, indicating the selected chunks.
The algorithm proceeds as follows. Initially, each chunk in $\mathbb{C}$ is evaluated using the state-of-the-art reranking model, resulting in a set of scores $\mathbb{S}$. Then, $\mathbb{C}$ is sorted in descending order based on $\mathbb{S}$. As a second step, we select the top $min\_k$ chunks as the initial $\mathbb{C}_s$, ensuring at least $min\_k$ chunks are chosen. Next, we examine the remaining chunks; if the score of a chunk exceeds $1/g$ times the score of its predecessor, we include it in $\mathbb{C}_s$. The selection process terminates when a chunk's score does not meet this condition, at which point $\mathbb{C}_s$ is returned.

\add{In short, our reranking method leverages a sophisticated, trainable scoring model and a dynamic selection process, enhancing contextual relevance for given questions.}

\section{LLM Self-Feedback}  \label{sec:feedback}

\subsection{Feedback Loop} \label{sec:feedback_loop}

As depicted in Figure~\ref{fig:overview} (C), after each QA session \scircled{3} conducted by the LLM, we organize a self-feedback prompt. This prompt integrates the question with the retrieved chunks, the generated answer, and any additional requests, as illustrated in Figure~\ref{fig:feedback}. Referred to as the “prompt of self-feedback," its purpose is to evaluate the LLM's current answer based on two criteria: (1) the answer's quality score to the question and (2) the suitability of the retrieved chunks - whether it contains redundant chunks or lacks necessary chunks for answering the question.
The feedback process yields two outcomes. The first is an evaluation score ranging from $1$ to $10$. The second denotes a context adjustment, signified as $-1$ or $1$. A context adjustment of $-1$ indicates the presence of redundant information within the retrieved chunks, while a $1$ suggests that additional information is required to answer the question sufficiently.

Upon receiving this feedback output, our initial step involves examining the score of answer quality. Should this score meet or exceed a threshold of feedback score $fs$, for example, $9$, the generated answer is considered acceptable and subsequently presented to the user. If not, the context adjustment is considered. $-1$ implies that the minimum number of retrieved chunks, $min\_k$, should be reduced by one, as outlined in Figure~\ref{fig:overview} (C) \scircled{6}($\dashrightarrow$). Conversely, $1$ requires an increase in $min\_k$ by one. Following any adjustments to $min\_k$, the process delineated in Figure~\ref{fig:overview} (B) \scircled{6}-\scircled{8} and Figure~\ref{fig:overview} (C) is repeated. This feedback loop continues until the feedback score surpasses the threshold or the feedback loop has been executed three times.

\noindent \textbf{Summary.} We propose to adjust the number of retrieved chunks using LLMs, further addressing the problems of noisy retrieval and missing retrieval.

\begin{table*}[!htb]
  \centering
  \caption{Effectiveness evaluation on NarrativeQA Dataset (using \texttt{GPT-4o-mini}).}
  \vspace{-.9em}
  \label{tab:narrativeqa accuracy}
  \setlength\tabcolsep{22pt}
      \begin{tabular}{|c||c|c|c|c|} \hline
      \diagbox{Model}{Metric}  &  \texttt{ROUGE}  &  \texttt{BLEU-1}   &  \texttt{BLEU-4}  &  \texttt{METEOR} \\ \hline
      \texttt{SBERT with \oursys}      &  \textbf{27.56\%}  &  \textbf{14.56\%}  &  \textbf{0.89\%}  &  \textbf{13.97\%}  \\ \hline
      \texttt{SBERT without \oursys}   &  27.34\%  &  12.99\%  &  0.94\%  & 12.61\%   \\ \hline
      \texttt{BM25 with \oursys}       &  \textbf{25.93\%}  &  \textbf{14.44\%}  &  \textbf{0.99\%}  &  \textbf{13.61\%}  \\ \hline
      \texttt{BM25 without \oursys}    &  22.12\%  &  10.95\%  &  0.89\%  &  11.07\%  \\ \hline
      \texttt{DPR with \oursys}        &  \textbf{24.67\%}  &  \textbf{12.15\%}  &  \textbf{0.99\%}  &  \textbf{11.75\%}  \\ \hline
      \texttt{DPR without \oursys}     &  22.94\%  &  10.87\%  &  0.25\%  &  11.12\%  \\ \hline
      \texttt{OpenAI Embedding with \oursys}       &  \textbf{26.56\%}  & \textbf{12.74\%}  &  \textbf{1.44\%}  &  \textbf{11.68\%}  \\ \hline
      \texttt{OpenAI Embedding without \oursys}    &  24.82\%  &  11.24\%  &  1.16\%  &  10.80\%  \\ \hline
      \end{tabular}
      \vspace{-.3em}
\end{table*}

\begin{table}[!htb]
  \centering
  \caption{Effectiveness evaluation on QuALITY and QASPER Dataset (using \texttt{GPT-4o-mini}).}
  \vspace{-.9em}
  \label{tab:quality and qasper accuracy with and without sfrag}
  \setlength\tabcolsep{12.6pt}
      \begin{tabular}{|c||c|c|} \hline
      \diagbox{Model}{Metric}  &  \makecell[c]{\texttt{Accuracy}\\ \texttt{(QuALITY)}}   &  \makecell[c]{\texttt{F1-Match}\\ \texttt{(QASPER)}} \\ \hline
      \texttt{SBERT with \oursys}    &  \textbf{73.14\%}  &  \textbf{40.23\%}  \\ \hline
      \texttt{SBERT without \oursys}   &  72.48\%  &  37.57\% \\ \hline
      \texttt{BM25 with \oursys}        &  \textbf{74.98\%}  &  \textbf{39.95\%}  \\ \hline
      \texttt{BM25 without \oursys}    &  72.18\%  &  37.30\%   \\ \hline
      \texttt{DPR with \oursys}         &  \textbf{74.37\%}  &  \textbf{40.06\%}   \\ \hline
      \texttt{DPR without \oursys}    &  72.38\%  &   37.41\%   \\ \hline
      \makecell[c]{\texttt{OpenAI Embedding}\\ \texttt{with \oursys}}         &  \textbf{78.30\%}  &  \textbf{41.23\%}  \\ \hline
      \makecell[c]{\texttt{OpenAI Embedding}\\ \texttt{without \oursys}}     &  75.32\%  &   38.94\%  \\ \hline
      \end{tabular}
\end{table}

\begin{table*}[!htb]
  \centering
  \caption{Ablation study on NarrativeQA Dataset (using \texttt{GPT-4o-mini}).}
  \vspace{-.9em}
  \label{tab:ablation on narrativeqa}
  \setlength\tabcolsep{24pt}
      \begin{tabular}{|c||c|c|c|c|} \hline
      \diagbox{Model}{Metric}  &  \texttt{ROUGE}  &  \texttt{BLEU-1}   &  \texttt{BLEU-4}  &  \texttt{METEOR} \\ \hline
      \texttt{Naive RAG}           &  28.45\%  &  12.73\%  &  0.29\%  &  12.73\%  \\ \hline
      \texttt{Naive RAG with Segmentation}  &  \textbf{29.74\%}  &  \textbf{13.98\%}  &  \textbf{0.33\%}  &  \textbf{12.84\%} \\ \hline
      \texttt{Naive RAG with Selection}  &  \textbf{29.15\%}  &  \textbf{13.18\%}  &  \textbf{0.42\%}  &  \textbf{12.92\%}  \\ \hline
      \texttt{Naive RAG with Feedback}   &  \textbf{30.59\%}  &  \textbf{14.89\%}  &  \textbf{0.81\%}  &  \textbf{14.34\%}  \\ \hline
      \oursys                      &  \textbf{31.65\%}  &  \textbf{15.27\%}  &  \textbf{1.70\%}  &  \textbf{14.42\%}  \\ \hline
      \end{tabular}
      \vspace{-.3em}
\end{table*}

\begin{table}[!htb]
  \centering
  \caption{Comparison on QASPER Dataset (using \texttt{GPT-3.5-turbo}).}
  \vspace{-.9em}
  \label{tab:comparision with baselines on qasper}
  \setlength\tabcolsep{2.5pt}
      \begin{tabular}{|c||c|c|} \hline
      \diagbox{Model}{Metric}  &  \texttt{GPT-3.5 F1-Match}  &  \texttt{GPT-4-o mini F1-Match} \\ \hline
      \makecell[c]{\texttt{Title+}\\ \texttt{Abstract}}    &  16.82\%  &  16.41\%   \\ \hline
      \texttt{BM25}   &  35.26\%  &  37.30\%   \\ \hline
      \texttt{DPR}    &  35.73\%  &  37.41\%   \\ \hline
      % \texttt{Raptor} &  53.1\%  &  55.7\%   \\ \hline
      \texttt{\oursys} &  \textbf{41.06\%}  &  \textbf{41.23\%}   \\ \hline
      \end{tabular}
      \vspace{-.3em}
\end{table}

\begin{table}[!htb]
  \centering
  \caption{Comparison on NarrativeQA dataset (using \texttt{UnifiedQA-3B}).}
  \vspace{-.9em}
  \label{tab:comparison with other baselines on narrativeqa}
  \setlength\tabcolsep{15pt}
      \begin{tabular}{|c||c|c|} \hline
      \diagbox{Model}{Metric}  &  \texttt{ROUGE}  &  \texttt{METEOR} \\ \hline
      \texttt{BiDAF}~\cite{kovcisky2018narrativeqa}       &  6.20\%   &  3.70\%   \\ \hline
      \texttt{BM25+BERT}~\cite{mou2020frustratingly}    &  15.5\%   &  5.0\%  \\ \hline
      \makecell[c]{\texttt{Recursively}\\ \texttt{Summarizing Books}}~\cite{wu2021recursively}   &  21.06\%   &  10.06\%  \\ \hline
      % \texttt{Retriever+Reader}~\cite{izacard2020distilling}   &  32.0\%  &  35.3\%  &  7.5\%  &  11.1\%   \\ \hline
      % \texttt{Raptor+UnifiedQA}    &  30.8\%  &  23.5\%  &  6.4\%  &  11.9\%   \\ \hline
      \texttt{\oursys+UnifiedQA}   &  \textbf{22.22\%}   &  \textbf{12.05\%}    \\ \hline
      \end{tabular}
      \vspace{-.3em}
\end{table}

\begin{table}[!htb]
  \centering
  \caption{Comparison on the QuALITY dataset (using \texttt{GPT-4}).}
  \vspace{-.9em}
  \label{tab:comparision with other baselines on quality2}
  \setlength\tabcolsep{2.7pt}
      \begin{tabular}{|c||c|c|} \hline
      \diagbox{Model}{Metric}  &  \makecell[c]{\texttt{Accuracy in}\\ \texttt{Test Set}}  &  \makecell[c]{\texttt{Accuracy in}\\ \texttt{Hard Set}} \\ \hline
      \texttt{Longformer-base}~\cite{beltagy2020longformer}    &  39.5\%  &  35.3\%    \\ \hline
      \texttt{DPR+DeBERTaV3-large}~\cite{pang2021quality}   &  55.4\%   &  36.1\%   \\ \hline
      \texttt{CoLISA(DeBERTaV3-large)}~\cite{dong2023colisa}    &  62.3\%  &  54.7\%    \\ \hline
      \texttt{RAPTOR+GPT-4}        &  82.6\%  &  76.2\%   \\ \hline
      \texttt{\oursys+GPT-4}    &   
      \textbf{90.10\%}  &  \textbf{76.3\%}    \\ \hline
      \end{tabular}
      \vspace{-.3em}
\end{table}

\begin{table*}[!htb]%(using \texttt{GPT4-o-mini})
  \centering
  \caption{\add{Memory usage, offline and online latency, and end-to-end performance of \our on a large scale \trivia dataset in a concurrent environment. (5x)/(10x) indicates five (time) times concurrency. Using \texttt{GPT4-o-mini}. }}
  \vspace{-.9em}
  \label{tab:revision_concurrency1}
  \setlength\tabcolsep{2.7pt} % Adjusts the column spacing
  \add{
  \begin{tabular}{|c|c|c|c|c|c|c|c|c|} \hline
    \textbf{Methods} & \makecell[c]{\textbf{Host memory} \\ \textbf{usage}} & \makecell[c]{\textbf{GPU memory} \\ \textbf{usage}} & \makecell[c]{\textbf{Latency of} \\ \textbf{building}\\ \textbf{vector database}} & \makecell[c]{\textbf{Latency of} \\ \textbf{segmentation}} & \makecell[c]{\textbf{Latency of} \\ \textbf{retrieval}} & \makecell[c]{\textbf{Latency of} \\ \textbf{feedback}} & \makecell[c]{\textbf{Latency of} \\ \textbf{answering}} & \textbf{F1-Match} \\ \hline
    Naive RAG & 0.580 GB & 0.000 GB & 0.000s & 9.696s (1066 tokens/s) & 0.914s & - & 1.817s & 0.704 \\ \hline
    BM25 + Naive RAG & 0.605 GB & 0.000 GB & 0.005s & 9.696s (1066 tokens/s) & 0.003s & - & 1.810s & 0.704 \\ \hline
    BM25 + SAGE & 4.870 GB & 2.200 GB & 0.005s & 5.070s (644 tokens/s) & 0.502s & 1.791s & 1.791s & 0.718 \\ \hline
    SAGE & 5.170 GB & 2.200 GB & 0.012s & 16.050s (725 tokens/s) & 0.8 + 0.50s & 1.831s & 1.794s & 0.724 \\ \hline
    SAGE (5x) & 6.320 GB & 2.765 GB & 0.012s & 16.050s (725 tokens/s) & 0.8 + 1.40s & 1.834s & 1.799s & 0.724 \\ \hline
    SAGE (10x) & 7.270 GB & 3.300 GB & 0.012s & 16.050s (725 tokens/s) & 0.8 + 2.25s & 1.831s & 1.791s & 0.724 \\ \hline
  \end{tabular}}
  \vspace{-.3em}
\end{table*}

\begin{table*}[!htb]
  \centering
  \caption{\add{Memory usage, offline and online latency, and end-to-end performance of \our on a large scale \trivia dataset in a concurrent environment. (5x)/(10x) indicates five (time) times concurrency. Using \texttt{UnifiedQA-3B}.}}
  \vspace{-.9em}
  \label{tab:revision_concurrency2}
  \setlength\tabcolsep{2.7pt} % Adjusts the column spacing
  \add{
  \begin{tabular}{|c|c|c|c|c|c|c|c|c|} \hline
    \textbf{Methods} & \makecell[c]{\textbf{Host memory} \\ \textbf{usage}} & \makecell[c]{\textbf{GPU memory} \\ \textbf{usage}} & \makecell[c]{\textbf{Latency of} \\ \textbf{building}\\ \textbf{vector database}} & \makecell[c]{\textbf{Latency of} \\ \textbf{segmentation}} & \makecell[c]{\textbf{Latency of} \\ \textbf{retrieval}} & \makecell[c]{\textbf{Latency of} \\ \textbf{feedback}} & \makecell[c]{\textbf{Latency of} \\ \textbf{answering}} & \textbf{F1-Match} \\ \hline
    Naive RAG & 3.256 GB & 12.20 GB & 0.004s & 9.696s (1066 tokens/s) & 0.914s & - & 1.089s & 0.652 \\ \hline
    BM25 + Naive RAG & 3.243 GB & 12.20 GB & 0.005s & 9.696s (1066 tokens/s) & 0.003s & - & 1.097s & 0.594 \\ \hline
    BM25 + SAGE & 13.150 GB & 15.24 GB & 0.005s & 5.070s (644 tokens/s) & 0.502s & 2.810s & 1.084s & 0.612 \\ \hline
    SAGE & 13.150 GB & 15.61 GB & 0.012s & 16.050s (725 tokens/s) & 0.8 + 0.50s & 2.793s & 1.091s & 0.671 \\ \hline
    SAGE (5x) & 15.500 GB & 16.175 GB & 0.012s & 16.050s (725 tokens/s) & 0.8 + 1.40s & 2.793s & 1.106s & 0.671 \\ \hline
    SAGE (10x) & 17.250 GB & 16.720 GB & 0.012s & 16.050s (725 tokens/s) & 0.8 + 2.25s & 2.801s & 1.097s & 0.671 \\ \hline
  \end{tabular}}
  \vspace{-.3em}
\end{table*}

\begin{table}[!htb]
  \centering
  \caption{\add{Verification of the effectiveness of feature augmentation for semantic segmentation on QASPER Dataset.}}
  \vspace{-.9em}
  \label{tab:revision_feature_aug}
  \setlength\tabcolsep{6pt}
  \add{
      \begin{tabular}{|c||c|} \hline
      \diagbox{\textbf{Features}}{\textbf{Metric}}  &  \texttt{Segmentation Accuracy}  \\ \hline
      $(x_1), (x_2)$ &  84.5\%    \\ \hline
      $(x_1), (x_2), (x_1-x_2)$ &  85.6\%    \\ \hline
      $(x_1), (x_2), (x_1*x_2)$ &  88.4\%   \\ \hline
      $(x_1), (x_2), (x_1-x_2), (x_1*x_2)$ &  \textbf{91.8\%}   \\ \hline
      \end{tabular}
      }
      \vspace{-.3em}
\end{table}

\begin{table}[!htb]
  \centering
  \caption{Comparison of cost efficiency (using \texttt{GPT-4o-mini}).}
  \vspace{-.9em}
  \label{tab:Cost Efficiency}
  \setlength\tabcolsep{6.1pt}
      \begin{tabular}{|c||c|c|c|} \hline
      \diagbox{Model}{Metric}  &  \makecell[c]{\texttt{Number}\\ \texttt{of tokens}} &  \texttt{Accuray}  &  \makecell[c]{\texttt{Relative Cost}\\ \texttt{Efficiency}} \\ \hline
      \texttt{BM25}   &  140699  &  65.0\%  &   0.646  \\ \hline
      \texttt{DPR}        &  142008  &  70.0\%  &  0.689    \\ \hline
      \texttt{SBERT}        &  140888  &  67.5\%  &  0.670    \\ \hline
      \texttt{\oursys}    &  \textbf{104939}  &  \textbf{75\%}  &  \textbf{1.0}   \\ \hline
      \end{tabular}
\end{table}

\section{Experiments}

\subsection{Experimental Setup}  \label{sec:exp_setup}

\noindent \textbf{\underline{Large language models.}} In our experiments, we utilize four LLMs in RAG frameworks, as detailed below.

\noindent(1) \texttt{GPT3.5 turbo}~\cite{brown2020language}. Developed by OpenAI, the GPT-3.5 Turbo model can comprehend and generate both natural language and code, marking a significant advancement in language model capabilities.

\noindent(2) \texttt{GPT4}~\cite{achiam2023gpt}. A successor to \texttt{GPT3.5 turbo}, GPT-4 is a large, multimodal model provided by OpenAI. It accepts both text and image inputs, generating text outputs with noteworthy accuracy and problem-solving capabilities surpassing those of its predecessor.

\noindent(2) \texttt{GPT4-o-mini}~\cite{gpt-4o-mini}. It is OpenAI's most advanced and cheapest LLM in the small models category. GPT-4o Mini is a multimodal model. It boasts superior intelligence to GPT-3.5 Turbo while matching its speed, emphasizing efficiency alongside enhanced cognitive performance.

\noindent(4) \texttt{UnifiedQA-3B}~\cite{khashabi2020unifiedqa}. Unified QA-3B focuses on question answering (QA) and uses a wide range of QA datasets. This helps it perform really well with different kinds of questions and subjects. It's a helpful tool for understanding language and creating responses.

\noindent \textbf{\underline{Datasets.}} We use three widely-used QA datasets: 
\noindent(1) QuALITY~\cite{pang2021quality}. 
This dataset comprises multiple-choice questions based on articles around 5,000 tokens each. It assesses reasoning across entire documents for QA tasks. It features a challenging portion, QuALITYHARD, with questions most annotators got wrong under time constraints. We present accuracies for both the full dataset and the QuALITYHARD subset. Performance is evaluated via the Accuracy metric.
\noindent(2) QASPER~\cite{dasigi2021dataset}. 
Spanning 5,049 questions from 1,585 NLP papers, QASPER delves into information within full texts. Answers vary, including answerable/unanswerable, yes/no, abstractive, and extractive. Performance is evaluated via the F1-Match metric.
\noindent(3) NarrativeQA~\cite{kovcisky2018narrativeqa, wu2021recursively}. 
Consisting of question-answer pairs from books and movie scripts (1,572 documents in total), the NarrativeQA demands a thorough grasp of the narrative for accurate responses, testing comprehension of extensive literary texts. Performance metrics include BLEU-1, BLEU-4, ROUGE, and METEOR.
\add{\noindent(4) Trival QA~\cite{2017arXivtriviaqa}. \trivia is a reading comprehension dataset containing over 650K question-answer-evidence triples, where passages with a maximum length of 470,719 tokens.}

\noindent \textbf{\underline{Metrics.}}
Our evaluation of RAG tasks encompasses two aspects: QA ability and \texttt{Cost-efficiency} (see Equation~\ref{equ:cost efficiency}). For assessing QA ability, we utilize different metrics tailored to each dataset:
For the QuALITY dataset, which features multiple-choice questions, we measure implementation success using \texttt{Accuracy}, meaning the ratio of correct multiple-choice questions to the total number.
In the QASPER and \add{Trival QA datasets}, we employ the F1-Match metric~\cite{rajpurkar2016squad}, considering its diverse question types.
For the NarrativeQA dataset, we apply a combination of metrics, including \texttt{ROUGE}~\cite{lin2004rouge}, \texttt{BLEU-1} and \texttt{BLEU-4}~\cite{papineni2002bleu}, and \texttt{METEOR}~\cite{banerjee2005meteor}, to evaluate comprehensive narrative understanding.

\noindent \textbf{\underline{Retrievers.}} We employ four distinct retrievers, most comprising an embedding model and a vector database, outlined as follows:
\noindent(1) \texttt{OpenAI Embedding}~\cite{openaiembeddings}. We use an OpenAI embedding model alongside a vector database to store and query embeddings. We specifically adopt the 'text-embedding-3-small'~\cite{openaiembeddings-small-3} model paired with a Faiss vector database~\cite{douze2024faiss}. 
\noindent(2) \texttt{BM25}~\cite{robertson1995okapi}. We use the BM25 algorithm, a probabilistic information retrieval model that ranks documents based on the term frequency-inverse document frequency (TF-IDF) of query terms appearing in each document, adjusted by the length of the document. This method is highly effective for text retrieval tasks, especially for shorter documents or passages.
\noindent(3) \texttt{DPR}~\cite{karpukhin2020dense}. We utilize the Dense Passage Retriever (DPR), which leverages a dense embedding model to encode passages and queries into the same vector space. The embeddings are then stored in a vector database, allowing for efficient and accurate retrieval based on vector similarity. This method has been shown to outperform traditional sparse retrieval techniques in many scenarios.
\noindent(4) \texttt{SBERT}~\cite{reimers2019sentence}. We employ the Sentence-BERT (SBERT) embedding model, which fine-tunes BERT using a Siamese network structure to generate semantically meaningful sentence embeddings. These embeddings are stored in a vector database, enabling quick and precise retrieval based on semantic similarity. SBERT is particularly effective for tasks requiring a nuanced understanding of sentence-level semantics.

\noindent \textbf{\underline{Comparison methods.}} In our evaluation, we compare a variety of methods as detailed below.
\noindent(1) \texttt{Naive RAG}. This approach divides continuous texts into segments of 200 tokens each, ensuring sentences are not split across different chunks. It then employs a LLM alongside the previously described retriever to execute the RAG task.
\noindent(2) \texttt{Title+Abstract}. This method utilizes the title and abstract of documents as the sole context for retrieval and generation.
\noindent(3) \texttt{BM25+BERT}~\cite{mou2020frustratingly}. This approach combines the BM25 algorithm for initial retrieval and BERT for re-ranking the retrieved documents based on their relevance to the question, leveraging the strengths of both sparse and dense retrieval methods.
\noindent(4) \texttt{Recursively Summarizing Books}~\cite{wu2021recursively}. This method involves recursively summarizing long documents into shorter, coherent summaries, which are then used as context for retrieval and generation tasks.
\noindent(6) \texttt{CoLISA}~\cite{dong2023colisa}. CoLISA first selects relevant sentences from a long passage according to the given question and its multiple options to construct a short context; then, it has multiple options that interact within a specific question in order to predict the final answer. We use the DeBERTaV3-large~\cite{he2021debertav3} language model in this method.
\noindent(7) \texttt{BiDAF}~\cite{kovcisky2018narrativeqa}. The BiDAF model employs a bi-directional attention flow mechanism to capture the interactions between the query and the context, enabling more precise QA by understanding the context at multiple levels.
\noindent(9) \texttt{Longformer-base}~\cite{beltagy2020longformer}. A model designed to process long sequenced data, addressing a limitation of the Transformer model. Longformer replaces the standard self-attention mechanism with a local windowed attention coupled with a task-specific global attention, which scales linearly with sequence length.
\noindent(9) \texttt{Raptor}~\cite{sarthi2024raptor}. This method innovates retrieval-augmented language models by creating a summarization tree for comprehensive document understanding.
\noindent(10) \oursys. Our RAG framework, which extends the \texttt{Naive RAG} method by incorporating the corpus segmentation technique as in Section~\ref{sec:segmentation}, the gradient-based chunks selection mechanism as in Section~\ref{sec:gradient selection}, and the self-feedback method as in Section~\ref{sec:feedback}. The default retriever used in \oursys is \texttt{OpenAI Embedding}.

\noindent \textbf{\underline{Hyper-parameters.}} The segmentation score threshold $ss$, referenced in Section~\ref{sec:segmentation inference}, is defined at 0.55. The length of coarse-grained $l$ discussed in Section~\ref{subsec:corpus segment} is set to $400$. The initial minimum number of retrieved chunks $min\_k$, as outlined in Section~\ref{subsec:gradient select}, is determined to be 7. The gradient threshold $g$ discussed in Section~\ref{subsec:gradient select} is set to 0.3. The feedback score threshold $fs$, discussed in Section~\ref{sec:feedback_loop}, is set to 9. 

\noindent \textbf{\underline{Environment.}} All experiments were performed on a ubuntu-22.04 server with a 20-core Intel(R) Xeon(R) 6242R 3.10GHz CPU, a Nvidia RTX3090 GPU, and 256GB DDR4 RAM.

\begin{figure}[!htb]
  \centering
  \includegraphics[width=0.45\textwidth]{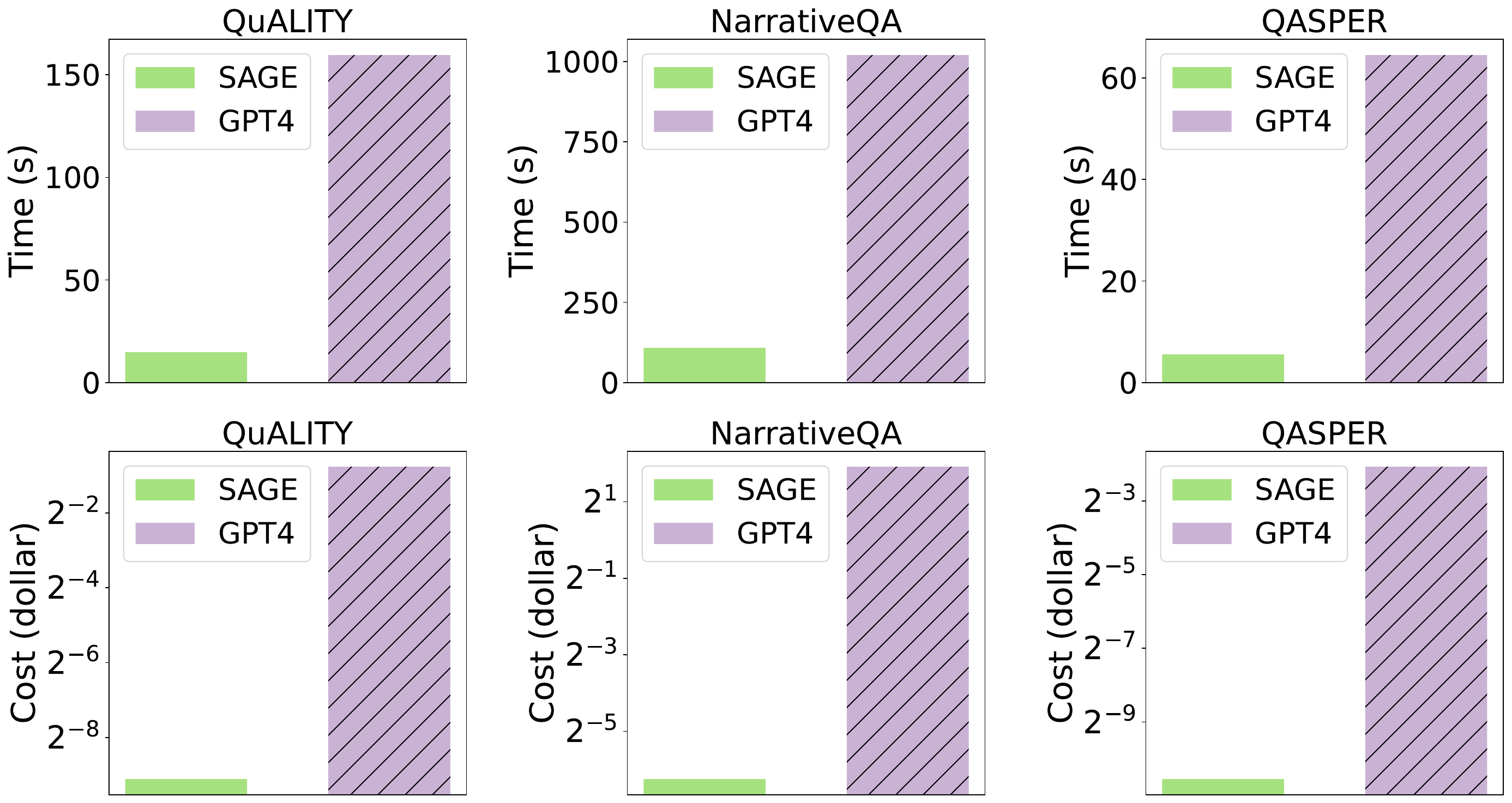}
  \vspace{-1em}
  \caption{Segmentation Overhead Evaluation.}
  \label{fig:partition_overhead}
  \vspace{-.2em}
\end{figure}

\subsection{End-to-End Question Answering Ability}
In this subsection, we evaluate the QA ability of various methods on different datasets.

\vspace{.2em}
\noindent $\blacktriangleright$ \textbf{Exp-1: Effectiveness on \texttt{NarrativeQA}.} \vspace{.15em}

We conduct an experiment on \texttt{NarrativeQA} dataset, to compare the End-to-End QA ability of different retrievers with and without the help of \oursys. The LLM used in this experiment is \texttt{GPT4-o-mini}. Table~\ref{tab:narrativeqa accuracy} shows the comparision results. We can observe that with the help of \oursys, the performance of retrievers \texttt{SBERT}, \texttt{BM25}, \texttt{DPR}, \texttt{OpenAI Embedding} increase 8.15\% in \texttt{ROUGE}, 17.27\% in \texttt{BLEU-1}, 81.51\% in \texttt{BLEU-4}, 11.89\% in \texttt{METEOR} on average. We can find that \oursys is effective for RAG systems with different retrievers. This is because \oursys is effective and orthometric with the embedding model and vector database modules in RAG systems.

\vspace{.2em}
\noindent $\blacktriangleright$ \textbf{Exp-2: Effectiveness on \texttt{QuALITY} and \texttt{QASPER}.} \vspace{.15em}

We conduct an experiment on \texttt{QuALITY} and \texttt{QASPER} datasets, to compare the End-to-End QA ability of different retrievers with and without the help of \oursys. The LLM used in this experiment is \texttt{GPT4-o-mini}. Table~\ref{tab:quality and qasper accuracy with and without sfrag} shows the comparison results. 
For \texttt{QuALITY} dataset, we can observe that the performance of retrievers \texttt{SBERT}, \texttt{BM25}, \texttt{DPR}, \texttt{OpenAI Embedding} increase 2.88\% in \texttt{Accuracy} on average. 
For \texttt{QASPER} dataset, we can observe that the performance of retrievers \texttt{SBERT}, \texttt{BM25}, \texttt{DPR}, \texttt{OpenAI Embedding} increase 6.79\% in \texttt{F1-Match} on average. The superior performance of RAG systems with \oursys is also because \oursys is effective and orthometric with the retrievers in RAG. We can find that the performance increment in \texttt{QuALITY} dataset is smaller than \texttt{QASPER} dataset. This is because \texttt{QuALITY} dataset is a multiple-choice QA dataset, and \texttt{Accuracy} in multiple-choice QA is easier than \texttt{F1-Match} in open-domain QA for both \oursys and the baselines. 

\vspace{.2em}
\noindent $\blacktriangleright$ \textbf{Exp-3: Comparison with baselines on \texttt{QuALITY}.} \vspace{.15em}

We conduct an experiment on \texttt{QuALITY} dataset, to compare the End-to-End QA ability of \oursys with other methods. Table~\ref{tab:comparision with other baselines on quality2} shows the comparision results between \oursys and (\texttt{Longformer-base}, \texttt{DPR+DeBERTaV3-large}, \texttt{CoLISA}, and \texttt{RAPTOR+GPT-4}). Note that \texttt{RAPTOR+GPT-4} is the state-of-the-art method in \texttt{QuALITY} dataset in the up-to-date leaderboard currently~\cite{quality_leadboard}. In the normal test set of \texttt{QuALITY} dataset, \oursys outperform these baselines by 128.4\%, 62.82\%, 44.78\%, and 9.2\%, respectively. In the hard test set of \texttt{QuALITY} dataset, \oursys outperforms these baselines by 116.1\%, 114.1\%, 39.49\%, and 0.13\%, respectively. These results verify the effectiveness of \oursys. We find that \oursys outperforms \texttt{RAPTOR+GPT-4} in the normal test set more than the hard test set. This is because many questions in the hard set are challenging to answer using RAG methods. For instance, there are some questions that need to be solved by reading the whole corpus and using the elimination method, which can not be solved by only retrieving a few information for LLMs.

\vspace{.2em}
\noindent $\blacktriangleright$ \textbf{Exp-4: Comparison with baselines on \texttt{NarrativeQA}.} \vspace{.15em}

We conduct an experiment on \texttt{NarrativeQA} dataset, to compare the End-to-End QA ability of \oursys with other methods. This experiment is conducted using \texttt{UnifiedQA-3B} language model. Table~\ref{tab:comparison with other baselines on narrativeqa} shows the comparison results between \oursys and (\texttt{BiDAF}, \texttt{BM25+BERT}, and \texttt{Recursively Summarizing Books}). We can find that \oursys outperforms these baselines by 258.4\%, 43.35\%, and 5.51\% in \texttt{ROUGE} and 225.7\%, 141\%, and 19.78\% in \texttt{METEOR}, respectively.

\vspace{.2em}
\noindent $\blacktriangleright$ \textbf{Exp-5: Comparison with baselines on \texttt{QASPER}.} \vspace{.15em}

We conduct an experiment on \texttt{QASPER} datasets to compare the End-to-End QA ability of \oursys with other methods. This experiment is conducted using \texttt{GPT-3.5 turbo} and \texttt{GPT4-o-min} language models. Table~\ref{tab:comparision with baselines on qasper} shows the comparison results between \oursys and (\texttt{Title+Abstract} and \texttt{Raptor}). For \texttt{GPT-3.5 turbo}, we find that \oursys outperforms these baselines by 144.1\%, 16.45\%, and 14.92\%, respectively. For \texttt{GPT4-o-mini}, we find that \oursys outperforms these baselines by 151.2\%, 10.54\%, and 10.21\%, respectively.

\add{\subsection{Scalability Evaluation}}  \label{exp:scalibility}

\vspace{.2em}
\noindent $\blacktriangleright$ \textbf{\add{Exp-6: Scalability evaluation.}}\vspace{.05em} 

\add{
We utilize the \trivia dataset to test the scalability of \our, under varying degrees of concurrency (5x and 10x). We conduct experiments using different language models: the GPT4-o-mini via a web interface for Table~\ref{tab:revision_concurrency1} and the UnifiedQA-3B on our local server for Table~\ref{tab:revision_concurrency2}. Our findings indicate that \our maintains superior performance over a naive RAG system even at high concurrency levels, with minimal increases in memory usage—only 27\% even at 10x concurrency. This efficiency is attributed to the high parallelism capabilities of GPUs, which allow a single LLM and segmentation model to handle multiple forward computations simultaneously.
Furthermore, the experiments demonstrate that while the retrieval latency increases slightly under higher concurrency (less than two seconds at 10x), the latency for feedback and answering processes remains consistent, regardless of the concurrency level. This stability is due to the effective parallel processing capacities of LLMs on GPUs, which effectively manage real-time latency demands.
Additionally, the construction of the vector database and segmentation processes perform only once, showing consistent latency across different concurrency levels, ensuring that the initial setup does not impact the system's overall responsiveness in subsequent operations.}

\subsection{Ablation Study}  \label{sec:exp_ablation}
In this subsection, we will verify the effectiveness of each main module of \oursys.

\vspace{.2em}
\noindent $\blacktriangleright$ \textbf{Exp-7: Ablation of each module of \oursys.}\vspace{.05em} 

We conduct an experiment on \texttt{NarrativeQA} dataset, to compare the End-to-End QA ability of \texttt{Naive RAG}, \texttt{Naive RAG} with each main module of \oursys, and \oursys. This experiment is conducted using \texttt{GPT4-o-min} language model. Table~\ref{tab:ablation on narrativeqa} shows the comparison results. We can find that \texttt{Naive RAG with Segmentation}, \texttt{Naive RAG with Selection}, \texttt{Naive RAG with Feedback}, and \texttt{SAGE} outperform \texttt{Naive RAG} by 4.53\%, 2.46\%, 7.52\% and 11.25\% in \texttt{ROUGE}, 9.82\%, 3.53\%, 16.97\% and 19.95\% in \texttt{BLEU-1}, 13.79\%, 44.83\%, 179.31\% and 486.21\% in \texttt{BLEU-4}, and 0.86\%, 1.49\%, 12.65\% and 13.28\% in \texttt{METEOR}. We find that the increments in \texttt{BLEU-4} and \texttt{METEOR} are smaller than that in \texttt{ROUGE} nad \texttt{BLEU-4}. These results verify the effectiveness of each module of \oursys. We find that \oursys outperforms \texttt{Naive RAG} with each main module. This is because the three modules do not negatively affect each other.

\vspace{.2em}
\noindent $\blacktriangleright$ \textbf{\add{Exp-8: Ablation of feature selection.}}\vspace{.05em}

\add{
To validate the effectiveness of the feature augmentation in training the segmentation model, we conduct an ablation study comparing the model's accuracy with and without the augmented features $x_1 - x_2$ and $x_1 \cdot x_2$. Specifically, we divided the articles from the QASPER dataset into a training set and a validation set using an 8:2 ratio. We then trained the segmentation model using various combinations of features on the training set and evaluated the segmentation accuracy on the validation set. As shown in Table~\ref{tab:revision_feature_aug}, we find that the segmentation performance is higher when using the features $(x_1), (x_2), (x_1-x_2), (x_1*x_2)$ compared to using only $(x_1), (x_2)$ or any other incomplete feature combinations.
}

\subsection{Cost Efficiency}  \label{sec:exp_cost}
In this subsection, we conduct an experiment to verify the superiority of \texttt{Cost-efficiency} of \oursys.

\vspace{.2em}
\noindent $\blacktriangleright$ \textbf{Exp-9: Comparison of Cost-efficiency.}\vspace{.05em} 

We conduct an experiment on \texttt{QuALITY} dataset, to compare the token consuming of a LLM, \texttt{Accuracy} of QA, and the \texttt{Cost-efficiency} of \oursys and naive rag systems with different embedding models. The LLM used in this experiment is \texttt{GPT4-o-mini}. Table~\ref{tab:Cost Efficiency} shows the results. We can observe that \oursys outperforms the three baselines by 53.85\%, 45.14\%, and 49.25\% of \texttt{Cost-efficiency} on average. Such improvements mainly because \oursys could achieve better QA performance while saving the input token of LLMs by eliminating noisy chunks.

\subsection{Segmentation Overhead and Cost} \label{exp:segmentation}
In this subsection, we conduct an experiment to compare the efficiency and money cost of corpus segmentation by our segmentation model and by a LLM.

\vspace{.2em}
\noindent $\blacktriangleright$ \textbf{Exp-10: Comparison of Segmentation Overhead.}\vspace{.05em} 

We conduct an experiment on three articles sampled from \texttt{QuALITY}, \texttt{NarrativeQA}, and \texttt{QASPER} datasets separately, to compare the time and money consuming of corpus segmentation by \oursys and \texttt{GPT-4} (See Section~\ref{sec:intro}). The time-consuming and money-consuming of our segmentation model is calculated by a rented server with a RTX3090 GPU, whose rental price is 5.3 dollars per day~\cite{vastai}. Table~\ref{fig:partition_overhead} shows that our segmentation model saves time by 90.71\% and money by 99.69\% than using \texttt{GPT-4} in \texttt{QuALITY} dataset, saves time by 89.43\% and money by 99.65\% than using \texttt{GPT-4} in \texttt{NarrativeQA} dataset, and saves time by 91.49\% and money by 99.72\% than using \texttt{GPT-4} in \texttt{QASPER} dataset. We can find that our segmentation method can save a huge amount of both time and money than using \texttt{GPT-4}. This is because our segmentation model is lightweight, achieving fast inference speed with high parallelism. Specifically, it can do inference fast and only occupies a little GPU memory (0.2 GB), allowing it to process a batch of 512 sentence pairs in one fast inference.

\begin{figure}[!t]
  \centering
  \includegraphics[width=0.475\textwidth]{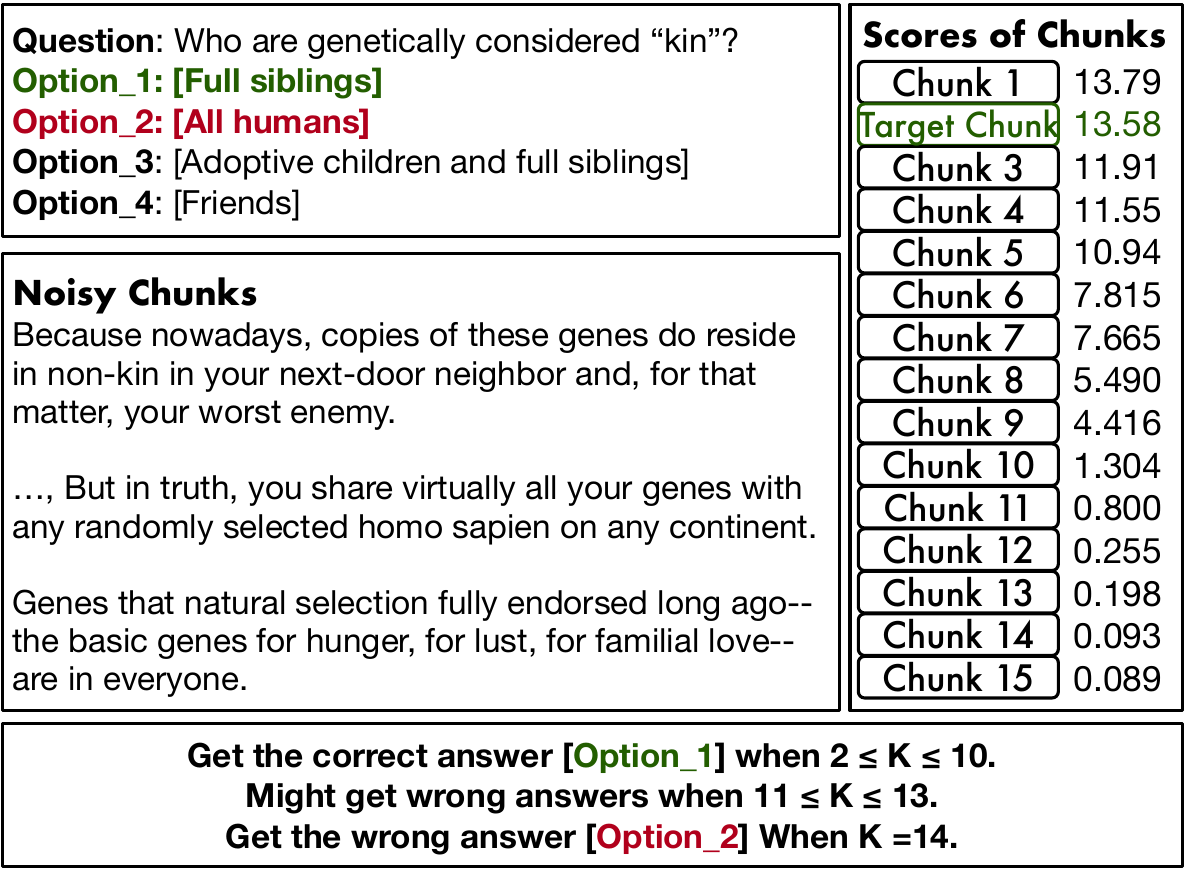}
  \vspace{-.5em}
  \caption{A case of \textit{noisy retrieval}.}
  % \vspace{-.3em}
  \label{fig:case1}
\end{figure}

\begin{figure}[!t]
  \centering
  \includegraphics[width=0.475\textwidth]{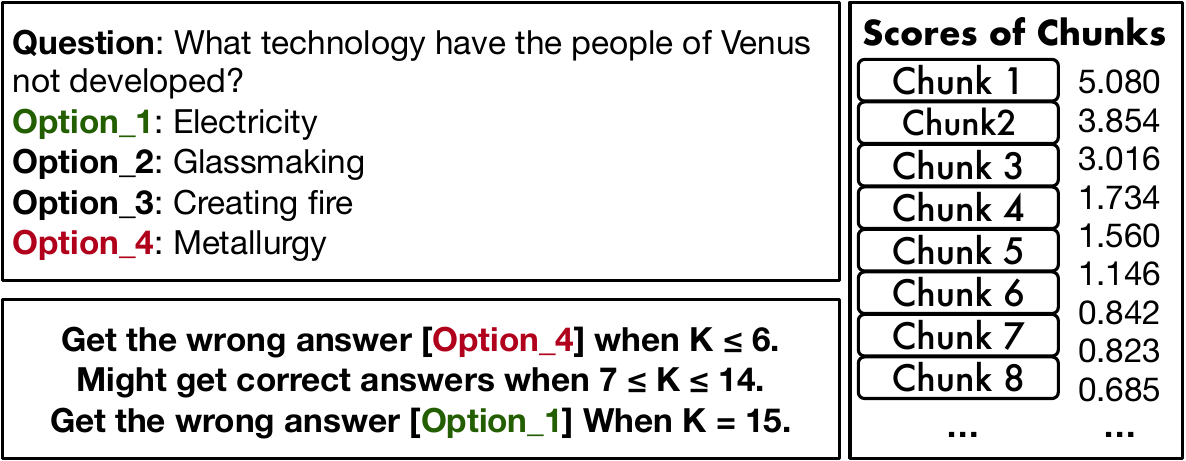}
  \vspace{-.5em}
  \caption{A case of \textit{missing retrieval}.}
  \label{fig:case3}
\end{figure}

\begin{figure}[!t]
  \centering
  \includegraphics[width=0.475\textwidth]{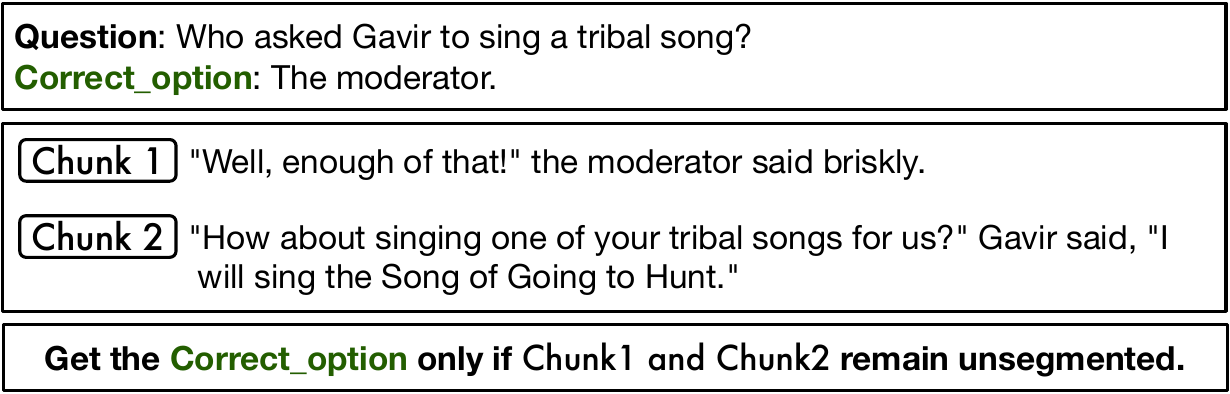}
  \vspace{-.5em}
  \caption{A case of \textit{ineffective corpus segmentation}.}
  \label{fig:case4}
\end{figure}

\begin{table}[!h]
    \centering
    \caption{Accuracy on QuALITY Dataset with different LLMs.}
    \vspace{-1.2em}
    \label{tab:comparison with baselines on quality}
    \setlength\tabcolsep{3.6pt}
        \begin{tabular}{|c||c|c|} \hline
        \diagbox{Model}{Metric}  &  \texttt{GPT-3.5 Accuracy}   &  \texttt{GPT-4o-mini Accuracy} \\ \hline
        \texttt{BM25}   &  62.70\%  &  73.50\%   \\ \hline
        \texttt{DPR}    &  60.4\%  &  73.0\%   \\ \hline
        % \texttt{Raptor} &  62.4\%  &  56.6\%   \\ \hline
        \texttt{\oursys}&    \textbf{64.5\%}    &  \textbf{77.1\%}   \\ \hline
        \end{tabular}
        % \vspace{-.2em}
\end{table}

\section{Insights of RAG Tasks} \label{sec:insight}

We summarize the following insights in RAG tasks:
\begin{enumerate}  % [label=\Roman*.]
    \item Noisy chunks retrieved considerably undermine the effectiveness of RAG systems (See \textbf{Exp-13}).
    \item Non-semantics-based corpus segmentation will impair the effectiveness of RAG systems (See \textbf{Exp-14}).
    \item The proficiency level of LLMs plays a crucial role in the effectiveness of RAG systems (See \textbf{Exp-15}).
    \item Interestingly, embedding models, though useful, are not as important as LLMs (See \textbf{Exp-15}).
\end{enumerate}

% First, xxx.

\vspace{.2em}
\noindent $\blacktriangleright$ \textbf{Exp-11: Case Study of noisy retrieval.}\vspace{.05em} 

Figure~\ref{fig:case1} demonstrates a real case from \texttt{QuALITY} dataset, illustrating the issue of noisy retrieval. For the given question, the correct answer is \texttt{Option1}, with the \texttt{Target Chunk} being ranked second in terms of relevance. However, there are some noisy chunks containing some information supporting the \texttt{Option2} in the other chunks. When with fewer noisy chunks ($2 \leq K \leq 10$), a LLM could choose the correct option. However, when with more noisy chunks ($K \ge  11$), the LLM might be misled into choosing the wrong answer \texttt{Option2}. 

\vspace{.2em}
\noindent $\blacktriangleright$ \textbf{Exp-12: Case Study of missing retrieval.}\vspace{.05em} 

Figure~\ref{fig:case3} demonstrates a real case from \texttt{QuALITY} dataset, showing the issue of missing retrieval. For the given question \textit{“Which technology not be developed?"}, it needs lots of context to use the elimination method to choose a correct answer. Therefore, we can find that $K < 6$ will lead to a wrong answer, and $K=15$ will lead to a correct answer. From the relevance scores of chunks, we can see that the sorted scores are smooth. With our gradient-based chunk selection algorithm, \oursys will choose more chunks and get the correct answer.

\vspace{.2em}
\noindent $\blacktriangleright$ \textbf{Exp-13: Case Study of incomplete chunks.} \vspace{.15em}

Figure~\ref{fig:case4} illustrates a case showing the issue of ineffective corpus segmentation. In the original corpus, \texttt{Chunk1} and \texttt{Chunk2} are consecutive. A LLM could produce a correct answer if consecutive (\texttt{Chunk1}, \texttt{Chunk2}) are retrieved. However, if using fixed-length segmentation, they may be segmented. Furthermore, the relevance score of \texttt{Chunk2} is higher than \texttt{Chunk1}, so they are impossible to retrieve together as in the original corpus. Finally, for the given question, it is infeasible to conclude that “The moderator asked Gavir to sing a tribal song.", resulting in a wrong answer.

\vspace{.2em}
\noindent $\blacktriangleright$ \textbf{Exp-14: Different LLMs and embedding models.} \vspace{.15em}

To evaluate the effectiveness of different LLMs, Table~\ref{tab:comparison with baselines on quality} shows the \texttt{Accuracy} of RAG methods using two proficiency levels of LLMs (\texttt{GPT-3.5-turbo} and \texttt{GPT-4o-mini}) on \texttt{QuALITY} dataset. The results show that RAG methods using \texttt{GPT-4o-mini} outperform RAG methods using \texttt{GPT-3.5-turbo} 17.38\%, 20.86\%, and 19.53\% on average. These results indicate the importance of the proficiency level of LLMs in RAG. To evaluate the effectiveness of different embedding models, Tabel~\ref{tab:narrativeqa accuracy} shows the performance of RAG methods with different embedding models. We can find that the performance order is $\texttt{SBERT} > \texttt{OpenAI Embedding} > \texttt{DPR} > \texttt{BM25}$, but the variance is smaller than that with different LLMs. This result suggests that while embedding models contribute to RAG's performance, they are not as influential as the choice of LLMs.
\section{Related Work}  \label{sec:related work}

\subsection{\add{Data Management}}
\add{Our work is a general framework to enhance the retrieval accuracy of information retrieval applications. It can be applied to the recent data management works that need to retrieve data according to the embedding distances of retrieved items. For instance, in multimodal retrieval, such as the systems developed by \cite{zheng2024knowledge} and \cite{li2024alleviating}, our framework could improve semantic alignment and cross-modal consistency, thereby enhancing overall retrieval precision.
For some efficient machine learning systems-related work, such as~\cite{zhangsageattention, zhang2024sageattention2,xi2024jetfire,wang2024efficient,li2024memory,zhang2025spargeattn,xi2025sparse,zhang2025sageattention2_wksp,zhang2025spargeattn_wksp,hu2025identifying,zhang2025accurate,zhang2025sageattention2++,zhang2025sageattention3,yang2025sparse}, our framework can be used in conjunction with these works within large models of RAG systems, thus ensuring both inference efficiency and retrieval accuracy. For some systems in databases, such as~\cite{UniBench, zhang2024pace, zhang2023autoce, zhang2024htap, pvldb/HyBench, sun2021learned}, our framework can be utilized to detach unstructured data from databases and use RAG technology for retrieval.
For distributed systems like those discussed in \cite{yao2023ragraph}, integrating our framework could optimize data retrieval processes, enhancing efficiency and reducing operational costs.
}

\subsection{Retrieval-Augmented Generation.}

The Retrieval-Augmented Generation (RAG) framework is a key advancement in natural language processing. It improves how generation systems understand and create text by using extra information from outside sources. RAG works by finding relevant information first and then using it to make better and more relevant text outputs.
For example, one study~\cite{lewis2020retrieval} introduced a model that efficiently locates pertinent documents to aid in answering questions or verifying facts. This method was much better at these tasks because it used external knowledge. Another important development, Dense Passage Retrieval (DPR)~\cite{karpukhin2020dense}, makes it easier to find the right pieces of information in large sets of data. This helps in creating accurate and to-the-point texts, which is the main goal of RAG.
Recent works have further refined RAG's effectiveness through various strategies, including prompt engineering~\cite{DBLP:journals/dase/ZhouSL24,DBLP:journals/pvldb/LiZZ24,sahoo2024systematic}, rewriting questions~\cite{gao2022precise,ma2023query}, adding knowledge basese~\cite{baek2023knowledge,DBLP:journals/pvldb/ZhouLSLCWLFZ24}, and iterative answering~\cite{cheng2024lift,shao2023enhancing,aop}. However, these methods often overlook the need to segment the corpus semantically and reduce irrelevant context retrieved, that \oursys, aims to solve.
In summary, RAG continues to evolve through continuous improvement in merging retrieval and generation. Our system \oursys focuses on the retrieval side.

\section{Conclusion and Future Work}  \label{sec:conclusion}
In this paper, we propose a framework of precise retrieval for RAG systems, named \oursys. We propose to train a segmentation model to segment a given corpus semantically and with low latency. Moreover, we propose a chunk selection algorithm to select the most relevant chunks rather than a larger fixed number of chunks. Lastly, we propose a self-feedback method to enable LLMs to adjust the number of retrieved chunks automatically. Experiments show that \oursys can achieve better QA performance with a lower cost of LLM inference than other baselines.

Looking ahead, we find three promising directions:
(1) \textbf{Multi-hop retrieval.} \add{We find the importance of scenarios where answers come from multiple documents like AiR Baleen~\cite{khattab2021baleen} and leave a comprehensive design for such scenarios as future work.}
(2) \textbf{Integration of \oursys with fine-tuning of LLMs.}
We have found that the proficiency level of LLMs is very important for the performance of a RAG system (See Section~\ref{sec:insight}). However, using the most powerful LLMs, e.g., \texttt{GPT-4}, is expensive. Fine-tuning is a simple way to enhance the QA ability of a LLM for a given corpus. For example, we can generate several batches of question-answer pairs to fine-tune \texttt{GPT-3.5-turbo}. Then, we might achieve the same QA performance based on the inexpensive LLM.
(3) \textbf{A more flexible chunk selection strategy.}
Currently, \oursys selects top chunks with higher relevance scores. Although \oursys selects a dynamic number of chunks, it is still possible there are useless chunks, e.g., the chunk with the highest relevance score is useless. Therefore, a more flexible chunk selection strategy might help. For example, we can train a LLM smart enough to select relevant chunks directly.

\newpage
\balance
\bibliographystyle{abbrv}
\bibliography{main}

\end{document}